\documentclass[preprint,12pt]{elsarticle}
\pdfoutput=1

\usepackage{graphicx}
\usepackage{algorithm}
\usepackage{algpseudocode}
\usepackage{epsfig}
\usepackage{amsfonts}
\usepackage{comment}
\usepackage{color}
\usepackage{amssymb}
\usepackage{amsmath, mathtools}
\usepackage{multirow}
\usepackage{etoolbox}
\usepackage{balance}
\usepackage{booktabs}
\usepackage{flushend}
\usepackage{lipsum}
\usepackage{epsf}
\usepackage{psfrag}
\usepackage{textcomp}
\usepackage{footnote}
\usepackage{tablefootnote}
\usepackage[flushleft]{threeparttable}
\usepackage{paralist}
\usepackage{mdwlist}
\usepackage{url}
\usepackage{verbatim}
\usepackage{alltt}
\usepackage{csquotes}
\usepackage{epstopdf}
\usepackage{float}
\usepackage{lineno}
\usepackage{slashbox}
\usepackage[table]{xcolor}
\usepackage{array}
\usepackage{boldline}
\usepackage{setspace}
\usepackage{mathrsfs}
\usepackage{stackengine}[2013-10-15]
\usepackage[T1]{fontenc}
\usepackage{frcursive}
\usepackage{calligra}
\usepackage{wedn}
\usepackage{float}
\usepackage{bm}
\usepackage{aurical}
\usepackage{upgreek}
\usepackage{stmaryrd}
\usepackage{enumerate}
\usepackage{adjustbox}

\usepackage{ dsfont }

\usepackage{ esint }
\usepackage{ tipa }

\usepackage[colorlinks = true,
    linkcolor = black,
    urlcolor  = black,
    citecolor = blue, 
    anchorcolor = magenta]{hyperref}

\definecolor{blue1}{RGB}{0,112,192}
\definecolor{bondiblue}{rgb}{0.0, 0.58, 0.71}
\definecolor{bostonuniversityred}{rgb}{0.8, 0.0, 0.0}

\DeclareMathOperator{\E}{\mathbb{E}}

\newcolumntype{?}{!{\vrule width 1pt}}
\newcolumntype{+}{!{\vrule width 1.25pt}}

\usepackage{tabularx}
\def\hlinewd#1{%
\noalign{\ifnum0=`}\fi\hrule \@height #1 %
\futurelet\reserved@a\@xhline}

\journal{Neurocomputing, Elsevier (Accepted)}

\begin{document}

\begin{frontmatter}

\title{Detecting Severity of Diabetic Retinopathy from Fundus Images: A Transformer Network-based Review}

\author[inst1]{Tejas Karkera}
\ead{tejkar10@gmail.com}
\affiliation[inst1]{organization={Northeastern University}, 
            city={Boston},
            postcode={02115}, 
            country={USA}}


\author[inst2]{Chandranath Adak\corref{cor1}}
\ead{adak32@gmail.com}
\cortext[cor1]{Corresponding author}
\affiliation[inst2]{organization={Dept. of CSE, IIT Patna},
            state={Bihar},
            postcode={801106}, 
            country={India}}
            
\author[inst3]{Soumi Chattopadhyay}
\ead{soumi61@gmail.com}
\affiliation[inst3]{organization={Dept. of CSE, IIT Indore},
            state={Madhya Pradesh},
            postcode={453552}, 
            country={India}}
            
\author[inst4]{Muhammad Saqib}
\ead{saqib.uet1@gmail.com}
\affiliation[inst4]{organization={Data61, CSIRO},
            state={NSW},
            postcode={2122}, 
            country={Australia}}

\begin{abstract}
Diabetic Retinopathy (DR) is considered one of the significant concerns worldwide, primarily due to its impact on causing vision loss among most people with diabetes. 
The severity of DR is typically comprehended manually by ophthalmologists from fundus photography-based retina images. 
This paper deals with an automated understanding of the severity stages of DR. In the literature, researchers have focused on this automation using traditional machine learning-based algorithms and convolutional architectures. However, the past works hardly focused on essential parts of the retinal image to improve the model performance. 
In this study, we adopt and fine-tune transformer-based learning models to capture the crucial features of retinal images for a more nuanced understanding of DR severity. 
Additionally, we explore the effectiveness of image transformers to infer the degree of DR severity from fundus photographs. 
For experiments, we utilized the publicly available APTOS-2019 blindness detection dataset, where the performances of the transformer-based models were quite encouraging. 
\end{abstract}


\end{frontmatter}


\section{Introduction}
\begin{figure}[!b]
\small
\centering
\begin{adjustbox}{width=\textwidth}
\begin{tabular}{ccccc}
\includegraphics[width=.0875\textwidth, height=.09\textwidth]{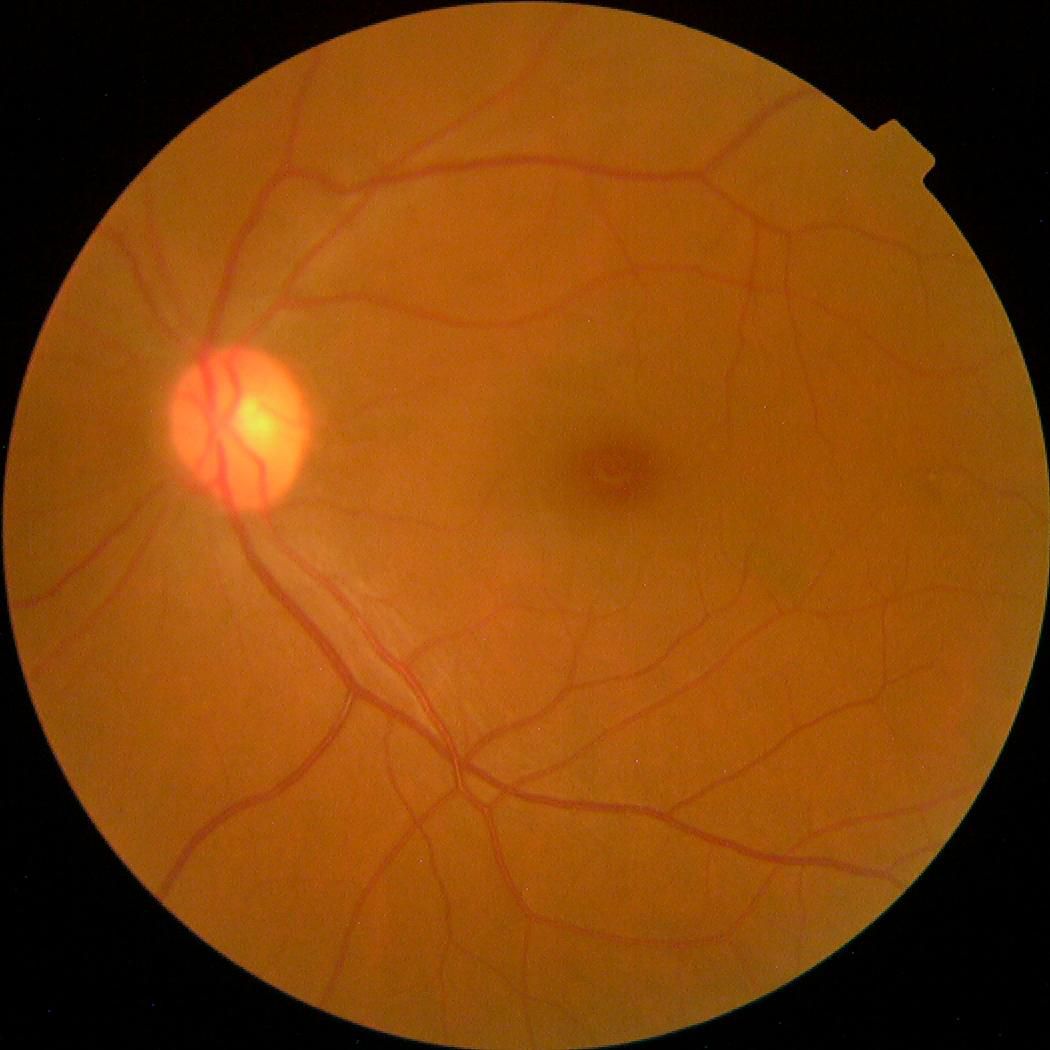}&
\includegraphics[width=.0875\textwidth, height=.09\textwidth]{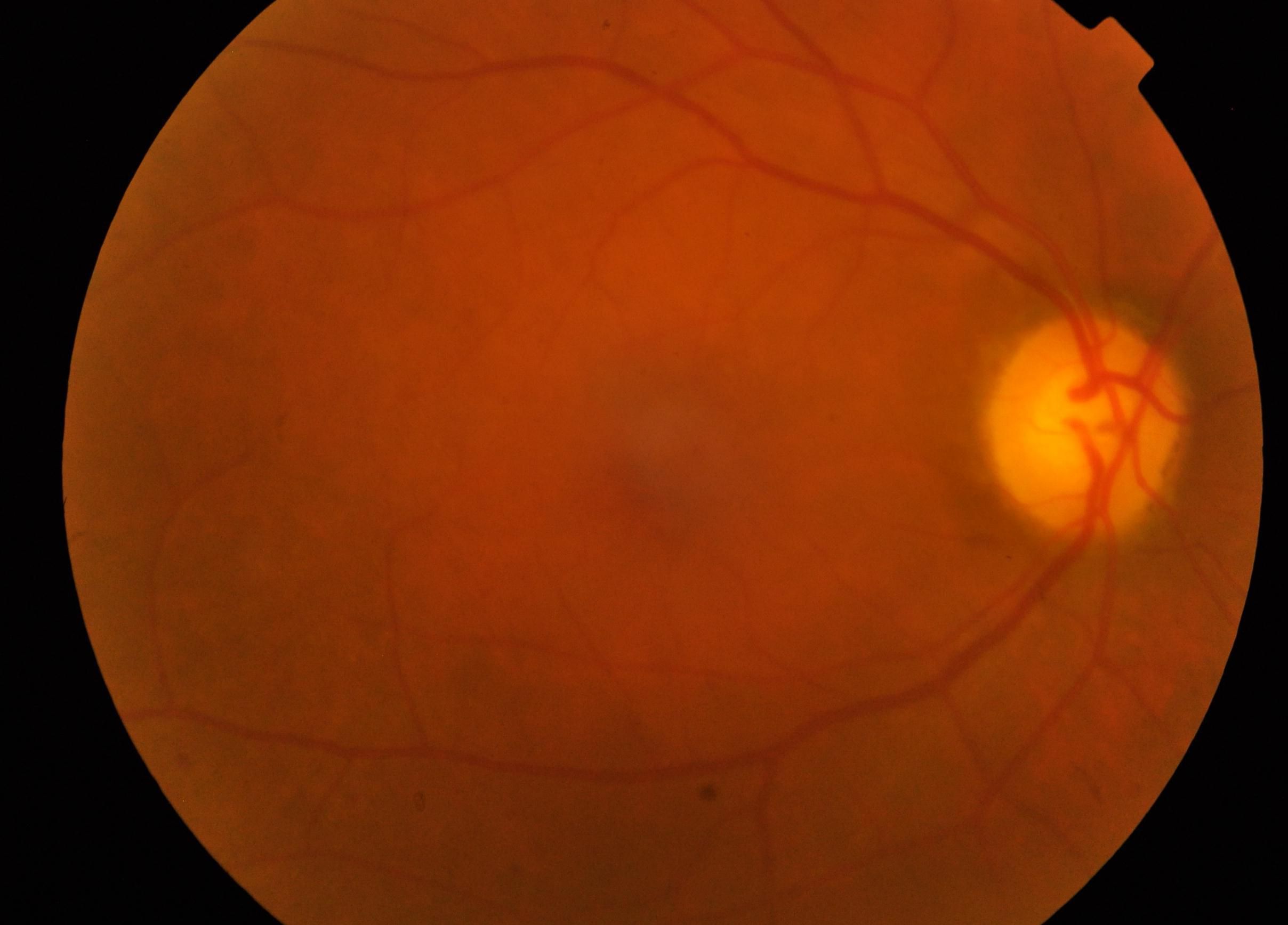}&
\includegraphics[width=.0875\textwidth, height=.09\textwidth]{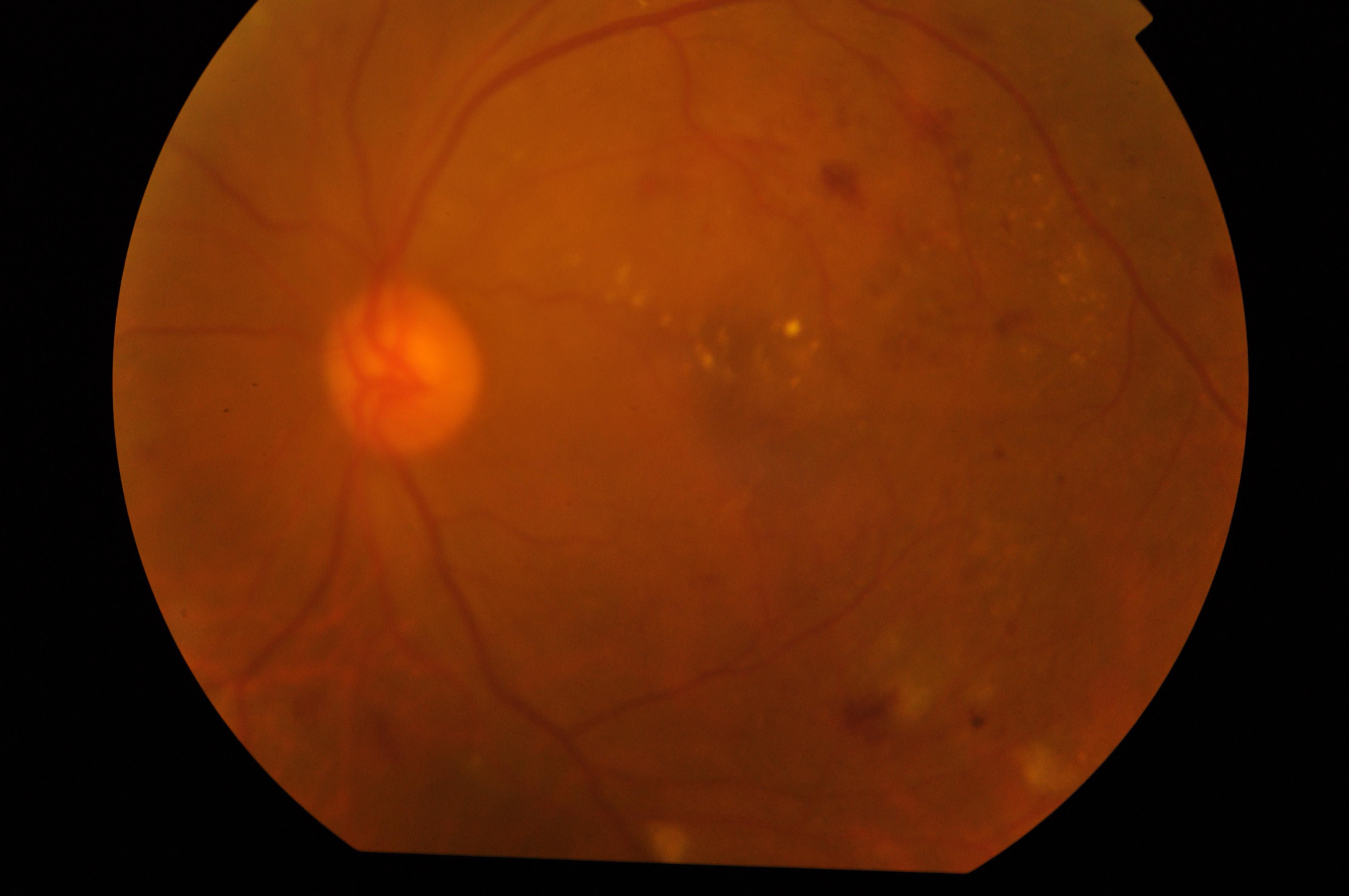}&
\includegraphics[width=.0875\textwidth, height=.09\textwidth]{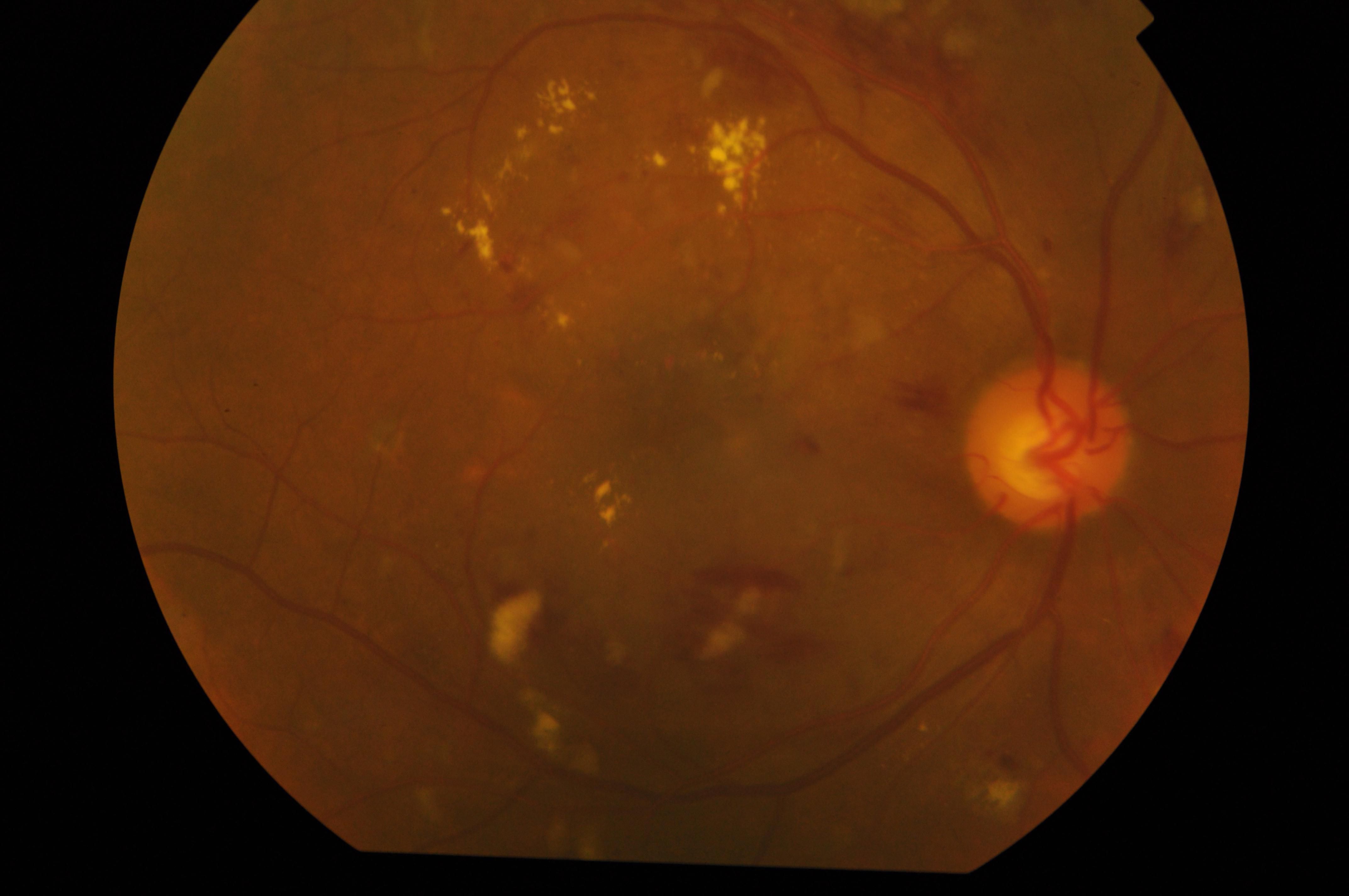}&
\includegraphics[width=.0875\textwidth, height=.09\textwidth]{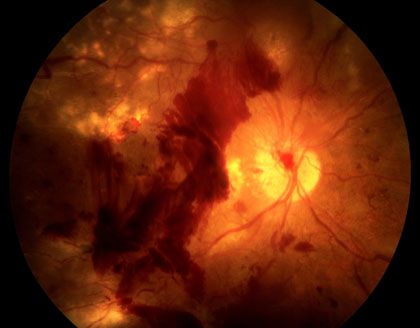}\\
negative & mild & moderate & severe & proliferative 
\end{tabular}
\end{adjustbox}
\caption{Fundus images with DR severity stages from APTOS-2019 \cite{aptos2019}
}
\label{fig:samples}
\end{figure}

\textcolor{black}{Diabetes Mellitus, commonly referred to as \emph{diabetes}, is a disorder where the patient experiences prolonged elevation in blood sugar levels. 
Diabetic Retinopathy (DR), a diabetes-related microvascular complication, involves retinal blood vessel damage, may lead to impaired vision and even blindness if left untreated \cite{survey2020,survey2019}. 
Studies estimated that around 99\% (or 60\%) of patients having type-I (or type-II) diabetes may develop DR within twenty years of diabetes onset \cite{survey2020}.
With a worldwide presence of DR patients of about 126.6 million in 2010, the current estimate is roughly around 191 million by 2030 \cite{zheng2012worldwide,jama_ting2017}. 
However, about 56\% of new DR cases can be reduced by timely treatment and monitoring of the severity \cite{Tymchenko}. Ophthalmologists analyze fundus images for lesion-based symptoms like microaneurysms, hard/ soft exudates, and hemorrhages to understand the severity stages of DR \cite{survey2020,survey2019}.
The positive DR is divided into the following stages \cite{Tymchenko}: 
($1$) \emph{mild} represents the earliest phase characterized by microaneurysms, 
($2$) \emph{moderate} signifies a stage where blood vessels start losing their transportation ability,  
($3$) \emph{severe} involves blockages in blood vessels, triggering the growth of new vessels,  
($4$) \emph{proliferative} denotes the advanced phase marked by the initiation of new blood vessel growth. 
Fig. \ref{fig:samples} shows some fundus images representing the DR severity stages. 
Manual assessment of fundus images for DR severity stage grading may yield inconsistencies due to a high patient volume, limited well-trained clinicians, prolonged diagnosis duration, ambiguous lesions, etc. 
Moreover, there may be disagreement among ophthalmologists in choosing the correct severity grade \cite{graderVariability}. 
Therefore, computer-aided techniques have come into the scenario for better diagnosis and broadening the prospects of early-stage detection \cite{survey2019}.}

\textcolor{black}{Automated DR severity stage detection from fundus photographs has evolved over the last two and half decades. Earlier, some image processing tools were utilized \cite{wavelet1,radon1}, but the machine learning (ML)-based DR approaches became popular in the early 21$^{st}$ century. The ML-based techniques mostly relied on hand-engineered features that were carefully extracted from the fundus images and then fed to a classifier, e.g., 
RF (Random Forest) \cite{RF3_casanova2014}, 
KNN (K-Nearest Neighbors) \cite{knn1_niemeijer2007}, 
SVM (Support Vector Machine) \cite{svm1_akram2013}, and 
ANN (Artificial Neural Network) \cite{ann1_usher2004}.
Although SVM and ANN-based models were admired in the DR community, the hand-engineered feature-based ML models require efficient prior feature extraction, which could lead to errors for complex fundus images \cite{survey2019,survey2020}. 
On the other hand, deep learning-based models extract features automatically through convolution operations \cite{bodapati2021deep,d2l}. 
Besides, since 2012, the surge of deep learning architectures in the computer vision community prominently influenced the DR severity analysis from fundus images \cite{survey2020}.
The past deep learning-based techniques mostly employed CNN (Convolutional Neural Network) \cite{cnn1_yu2017,survey2020}. 
However, the ability to give attention to specific regions/features and fade the remaining portions hardly exists in classical CNNs. 
Also, CNN-based classifiers may compromise spatial relationships among the learned features. To overcome the limitations of CNNs, capsule networks have been applied \cite{capsule2}, or classifications have been performed by incorporating attention mechanisms \cite{zoomin_wang2017,farag2022automatic}.
Although multiple studies exist in the literature \cite{survey2020,survey2019} and efforts were made to detect the existence of DR in the initial stages of its development, there is still room for improving the performance by incorporating higher degrees of automated feature extraction using more sophisticated deep learning models.}

\textcolor{black}{In this paper, we employ and fine-tune the transformer model for leveraging its MSA (Multi-head Self-Attention) \cite{vit} to focus on the specific region in fundus images that reveal signs of DR severity. 
Moreover, transformer models have demonstrated high performance in recent days for various computer vision tasks \cite{bracsoveanu2020visualizing,vit}. 
Initially, we adopted ViT (Visual Transformer) \cite{vit} for detecting DR severity due to its superior performance in image classification tasks. 
ViT dissects the input image into a sequence of patches and applies global attention \cite{vit}. 
Since standard ViT requires hefty amounts of data, we also explored some other image transformer models, such as 
CaiT (Class-attention in image Transformers) \cite{cait}, 
DeiT (Data-efficient image Transformer) \cite{deit}, and 
BEiT (Bidirectional Encoder representation for image Transformer) \cite{beit}. 
CaiT is a modified version of ViT and employs specific class-attention \cite{cait}. 
DeiT uses knowledge distillation, which transfers the knowledge from one network to another and builds a teacher-student hierarchical network \cite{deit}. 
BEiT draws inspiration from BERT (Bidirectional Encoder Representations from Transformers) \cite{devlin2018bert} to implement masking of image patches and to model the same for pre-training the ViT \cite{beit}. 
For experiments, we used the publicly available APTOS-2019 blindness detection dataset \cite{aptos2019}, where individual image transformers did not perform well. 
Therefore, we ensembled the above fine-tuned image transformers to seek better predictive performance. The ensembled image transformer obtained quite encouraging results for DR severity stage detection. 
This is one of the earliest attempts to adopt and ensemble image transformers for DR severity stage detection, which is the main \textbf{contribution} of this paper.}

The rest of the paper is organized as follows. 
Section \ref{related} discusses the relevant literature about DR and Section \ref{methodology} presents the proposed methodology. 
Then Section \ref{experimental_result} analyzes and discusses the experimental results. 
Finally, Section \ref{conc} concludes this paper.

\section{Related Work}\label{related}

This section briefly presents the literature on DR severity detection from fundus images. 
The modern grading of DR severity stages can be traced in the report by the ETDRS research group \cite{grading_ETDRS}. 
In the past, some image processing-based (e.g., wavelet transform\cite{wavelet1}, radon transform \cite{radon1}) strategies were published. 
For the last two decades, machine learning and deep learning-based approaches have shown dominance. 
We broadly categorize the related works into 
(a) hand-engineered feature-based models \cite{RF1_acharya2017,knn1_niemeijer2007,svm2_adal2017}, and 
(b) deep feature-based models \cite{survey2019}, which are discussed below.

\subsection{Hand-engineered Feature-based Models}
The hand-engineered feature-based models mostly employed RF \cite{RF1_acharya2017}, KNN \cite{knn2_Tang2013}, SVM \cite{svm2_adal2017}, ANN \cite{ann3_herliana2018} for detecting DR severity stages.
%
Acharya et al. \cite{RF1_acharya2017} employed a decision tree with discrete wavelet/cosine transform-based features extracted from retinal images.
Casanova et al. \cite{RF3_casanova2014} introduced RF for DR severity stage classification. 
In \cite{RF2_sanroma2016}, RF was also used to assess DR risk. 
%
KNN classifier was employed in \cite{knn1_niemeijer2007} to detect drusen, exudates, and cotton-wool spots for diagnosing DR.
Tang et al. \cite{knn2_Tang2013} used KNN for retinal hemorrhage detection from fundus photographs.
%
In \cite{svm2_adal2017}, retinal changes due to DR was detected by using SVM. 
Akram et al. \cite{svm1_akram2013} used SVM and GMM (Gaussian Mixture Model) with enhanced features such as shape, intensity, and statistics of the affected region to identify microaneurysms for early detection of DR.
%
ANN was employed in \cite{ann1_usher2004} to classify lesions for detecting DR severity. 
Osareh et al. \cite{ann2_osareh2009} employed fuzzy C-means-based segmentation and genetic algorithm-based feature selection with ANN to detect exudates in DR.
In \cite{ann3_herliana2018}, particle swarm optimization was used for feature selection, followed by ANN-based DR severity classification.

\subsection{Deep Feature-based Models}
The past deep architectures mostly used CNN for tackling DR severity.
For example, Yu et al. \cite{cnn1_yu2017} used CNN for detecting exudates in DR, 
Chudzik et al. \cite{cnn2_chudzik2018} worked on microaneurysm detection using CNN with transfer learning and layer freezing, 
Gargeya and Leng \cite{cnn3_gargeya2017} employed CNN-based deep residual learning to identify fundus images with DR.
In \cite{jama_ting2017}, CNN was also used to identify DR severity stages and some related eye diseases, e.g., glaucoma and AMD (Age-related Macular Degeneration). 
In \cite{cnn4_wan2018}, some classical CNN architectures (e.g., AlexNet, VGG Net, GoogLeNet, ResNet) were employed for DR severity stage detection.
Wang et al. \cite{zoomin_wang2017} proposed Zoom-in-Net that combined CNN, attention mechanism, and a greedy algorithm to zoom in the region of interest for handling DR.
A modified DenseNet169 architecture in conjunction with the attention mechanism was used in \cite{farag2022automatic} to extract refined features for DR severity grading.
In \cite{kassani2019}, a modified Xception architecture was employed for DR classification. 
TAN (Texture Attention Network) was proposed in \cite{alahmadi2022texture} by leveraging style (texture features) and content (semantic and contextual features) re-calibration mechanism. 
Tymchenko et al. \cite{Tymchenko} ensembled three CNN architectures (EfficientNet-B4 \cite{efficientnet}, EfficientNet-B5, and SE- ResNeXt50 \cite{squeeze}) for DR severity detection.
Very recently, a few transformer-based models have come out, e.g., CoT-XNet \cite{cotxnet} that combined contextual transformer and Xception architecture, SSiT \cite{ssit} that employed self-supervised image transformers guided by saliency maps.

\section{Methodology}
\label{methodology}
This section first formalizes the problem statement, which is then followed by the proposal of solution architecture.

\subsection{Problem Formulation}
\label{subsec:problemFormulation}
In this work, we are given an image $I$ captured by the fundus photography, which is input to the architecture.
The task is to predict the severity stage of DR among negative, mild, moderate, severe, and proliferative, from $I$.
We formulate the task as a multi-class classification problem \cite{d2l}. Here, from $I$, features are extracted and fed to a classifier to predict the DR severity class labels \texthtc, where \texthtc~$ = \{0,1,2,3,4\}$ corresponds to \{negative, mild, moderate, severe, proliferative\}, respectively.

\subsection{Solution Architecture}
To detect the severity stage of DR from a fundus photograph, we adopt image transformers, i.e.,  
ViT \cite{vit}, 
BEiT \cite{beit}, 
CaiT \cite{cait}, and 
DeiT \cite{deit}, 
and ensemble them. 
However, we preprocess raw fundus images before feeding them into the transformers, which we discuss first.

\subsubsection{Preprocessing}
\label{Preprocessing}
The performance of deep learning models is susceptible to the quality and quantity of data being passed to the model. Raw data as input can barely account for the best achievable performance of the model due to possible pre-existing noise and inconsistency in the images.
Therefore, a definite flow of preprocessing is essential to train the model better \cite{d2l}.

We now discuss various preprocessing and augmentation techniques \cite{goodfellow2016deep,d2l} applied to the raw fundus photographs for better learning. 
In a dataset, the fundus images may be of various sizes; therefore, we resize the image $I$ into $256 \times 256$ sized image $I_z$.  
\textcolor{black}{Deep networks are data-hungry, and several augmentations have been applied to address diverse issues associated with deep networks \cite{augment1}, including model overfitting, and to enhance the robustness of the models. Therefore, we perform data augmentations on the training set (DB$_{tr}$), where we use centre cropping with central\_fraction = 0.5, horizontal/vertical flip, random rotations within a range of $[0^o,45^o]$, random brightness-change with max\_delta = 0.95, 
random contrast-change in the interval $[0.1, 0.9]$. We also apply CLAHE (Contrast Limited Adaptive Histogram Equalization) \cite{clahe} on 30\% samples of DB$_{tr}$, which ensures over-amplification of contrast in a smaller region instead of the entire image.} 
\textcolor{black}{Intensities in medical images are usually inhomogeneous and may affect the performance of the automated image analysis methods. 
In the literature, some normalization methods have been implemented with different image types to obtain good performance \cite{inorm1}. 
Also, image noise may affect computerized methods, and different denoising algorithms have been applied to different types of images \cite{inorm3}. 
However, they may lead to increased computational costs. Therefore, the proposed approach has been refrained from applying them.}

\subsubsection{Transformer Networks}
\label{Transformer Models}
Deep learning models in computer vision tasks have long been dominated by CNN to extract high-level feature maps by passing the image through a series of convolution operations before feeding into the MLP (Multi-Layer Perceptron) for classification \cite{sarvamangala2021convolutional}.
In recent days, transformer models have shown a substantial rise in the NLP (Natural Language Processing) domain due to its higher performances \cite{bracsoveanu2020visualizing}. 
\textcolor{black}{In a similar quest to leverage high-level performance through transformers, it has been introduced in image classification and some other computer vision-oriented tasks \cite{vit}. Moreover, the transformer model has lesser image-specific inductive bias than CNN \cite{vit}.}  

To identify the severity stages of DR from fundus images, here we efficiently adopt and ensemble some image transformers, e.g., 
ViT \cite{vit}, 
BEiT \cite{beit}, 
CaiT \cite{cait}, and 
DeiT \cite{deit}. 
Before focusing on our ensembled transformer model, we discuss the adaptation of individual image transformers for our task, and start with ViT.  

\emph{{\ref{Transformer Models}.1. Vision Transformer (ViT)}:}
\textcolor{black}{The ViT model adopts the idea of text-based transformer models \cite{nlp_transformers}, where the idea is to take the input image as a series of image patches instead of textual words, and then extract features to feed it into an MLP \cite{vit}.} 

\begin{figure}[!b]
\centering
\includegraphics[width=0.75\linewidth]{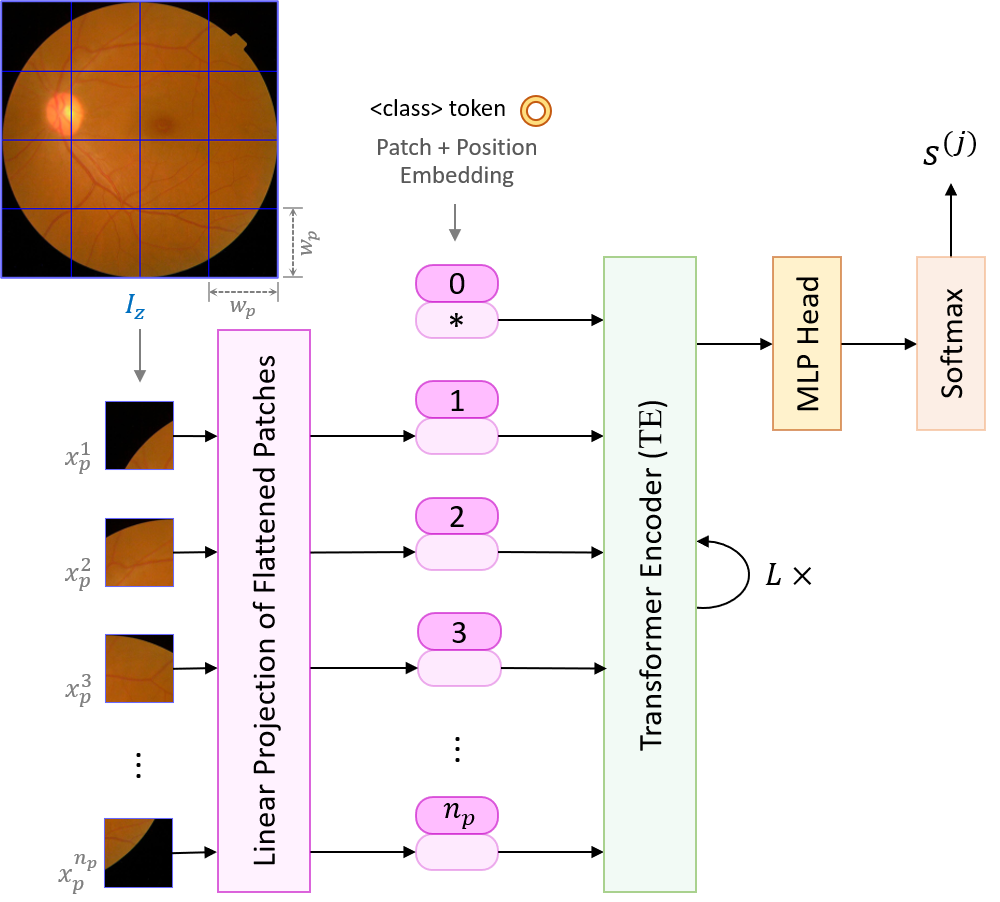}
\caption{Workflow of ViT}
\label{fig:vit}
\end{figure}

The pictorial representation of ViT is presented in Fig. \ref{fig:vit}.
Here, the input image $I_z$ is converted into a sequence of flattened patches $x_p^i$ (for $i=1,2,\dots,n_p$), each with size $w_p \times w_p \times c_p$, 
where $c_p$ denotes the number of channels of $I_z$. Here, $c_p=3$, since $I_z$ is an RGB fundus image. In our task, $I_z$ is of size $256 \times 256$, and empirically, we choose $w_p=64$, which results $n_p = (\frac{256}{64})^2 = 16 $. 
Each patch $x_p^i$ is flattened further and mapped to a $D$-dimensional latent vector (i.e., patch embedding $z_0$) through transformer layers using a trainable linear projection, as below.
\begin{equation}
    z_0 = [ x_{class}~;~ x_p^1\E~;~ x_p^2\E~;~ \dots~;~ x_p^{n_p}\E ] +\E_{pos}
\end{equation}
where, $\E$ is the patch embedding projection, 
$ \E \in \mathbb{R}^{w_p\times w_p\times C \times D}$; 
$\E_{pos}$ is the position embeddings added to patch embeddings to preserve the positional information of patches, 
$ \E_{pos} \in \mathbb{R} ^{(n_p+1) \times D} $; 
$x_{class} = z_0^0$ is a learnable embedding \cite{devlin2018bert}.

\begin{figure}
\centering
\includegraphics[width=0.9\linewidth]{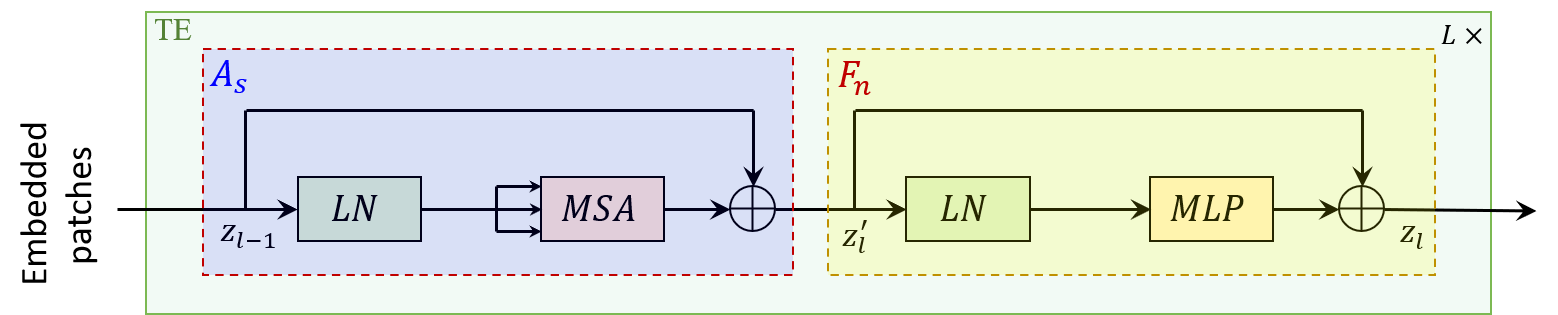}
\caption{Internal view of a transformer encoder (TE)}
\label{fig:encoder}
\end{figure}

After mapping patch images to the embedding space with positional information, we add a sequence of transformer encoders \cite{vaswani2017attention,vit}. 
The internal view of a transformer encoder can be seen in Fig. \ref{fig:encoder}, 
which includes two blocks $A_s$ and $F_n$. The $A_s$ and $F_n$ contain $MSA$ (Multi-head Self-Attention) \cite{vit} and $MLP$ \cite{d2l} modules, respectively. $LN$ (Layer Normalization) \cite{ln} and residual connection \cite{d2l} are employed before and after each of these modules, respectively. 
This is shown in equation \ref{eq2} with general semantics. 
Here, the $MLP$ module comprises two layers having $4D$ and $D$ neurons with GELU (Gaussian Error Linear Unit) non-linear activation function similar to \cite{vit}.
\begin{equation} \label{eq2}
\begin{split}
    z_l' = MSA(LN(z_{l-1})) + z_{l-1}; \\
    z_l = MLP(LN(z_l')) + z_l';~ 
    l=1,2, \dots ,L
\end{split}
\end{equation}
where, $L$ is the total number of transformer blocks. 
%
The core component of the transformer encoder is $MSA$ with $h$ heads, where each head includes $SA$ (Scaled dot-product Attention) \cite{vit,vaswani2017attention}. 
Each head $i \in \{1,2,...,h\}$ of $MSA$ calculates a tuple comprising query, key, and value \cite{vit}, i.e., ($Q^i, K^i, V^i$) as follows.
\begin{equation}
Q^i = X W_Q^i ~;~
K^i = X W_K^i ~;~
V^i = X W_V^i
\end{equation}
where, $X$ is the input embedding, and $W_Q$, $W_K$, $W_V$ are the weight matrices used in the linear transformation. 
The tuple ($Q,K,V$) is fed to $SA$ that computes the attention required to pay to the input image patches, as below.
\begin{equation}
    SA(Q,K,V) = \psi \left(\frac{QK^T}{\sqrt{D_h}}\right)V 
\end{equation}
where, $\psi$ is softmax function, and $D_h = D/h$. 
The outcomes of $SA$s across all heads are concatenated in $MSA$, as follows.
\begin{equation}
MSA(Q,K,V) = [SA^1~;~ SA^2~;~ \dots ~;~ SA^h]W_L
\end{equation}
where $W_L$ is a weight matrix. 

After multiple transformer encoder blocks, the <class> token \cite{devlin2018bert} enriches with the contextual information. The state of the learnable embedding at the outcome of the Transformer encoder ($z_L^0$) acts as the image representation $y$ \cite{vit}.
\begin{equation}
    y = LN(z_L^0)
\end{equation}

Now, as shown in Fig. \ref{fig:vit}, we add an MLP head containing a hidden layer with 128 neurons. To capture the non-linearity, we use Mish \cite{mish} here. In the output layer, we keep five neurons with softmax activation function to obtain probability distribution $s^{(j)}$ in order to classify a fundus photograph into the abovementioned five severity stages of DR.


\emph{{\ref{Transformer Models}.2. Data efficient image Transformers (DeiT)}:}
\textcolor{black}{For a lower amount of training data, ViT does not generalize well. In this scenario, DeiT can perform reasonably well and uses lower memory \cite{deit}. 
DeiT adopts the ViT-specific strategy and merges with the teacher-student scheme through knowledge distillation \cite{TeacherStud}. 
The crux of DeiT is the knowledge distillation mechanism, which is basically the knowledge transfer from one model (teacher) to another (student) \cite{deit}.}
Here, we use EfficientNet-B5 \cite{efficientnet} as a teacher model that is trained apriori. 
The student model uses a transformer, which learns from the outcome of the teacher model through attention depending on a distillation token \cite{deit}. 
In this work, we employ hard-label distillation \cite{deit}, where the hard decision of the teacher is considered as a true label, i.e., $y_t = \text{argmax}_c Z_t(c)$.
The hard-label distillation objective is defined as follows. 
\begin{equation}
    {\mathcal{L}}_{global}^{hard} = 0.5~{\mathcal{L}}_{CE}(\psi(Z_s),y) + 0.5~{\mathcal{L}}_{CE}(\psi(Z_s),y_t)
\end{equation}
where, ${\mathcal{L}}_{CE}$ is the cross-entropy loss on ground-truth labels $y$, $\psi$ is the softmax function, $Z_s$ and $Z_t$ are the student and teacher models' logits, respectively. 
Using label smoothing, hard labels can be converted into soft ones \cite{deit}.

\begin{figure}
\centering
\includegraphics[width=0.7\linewidth]{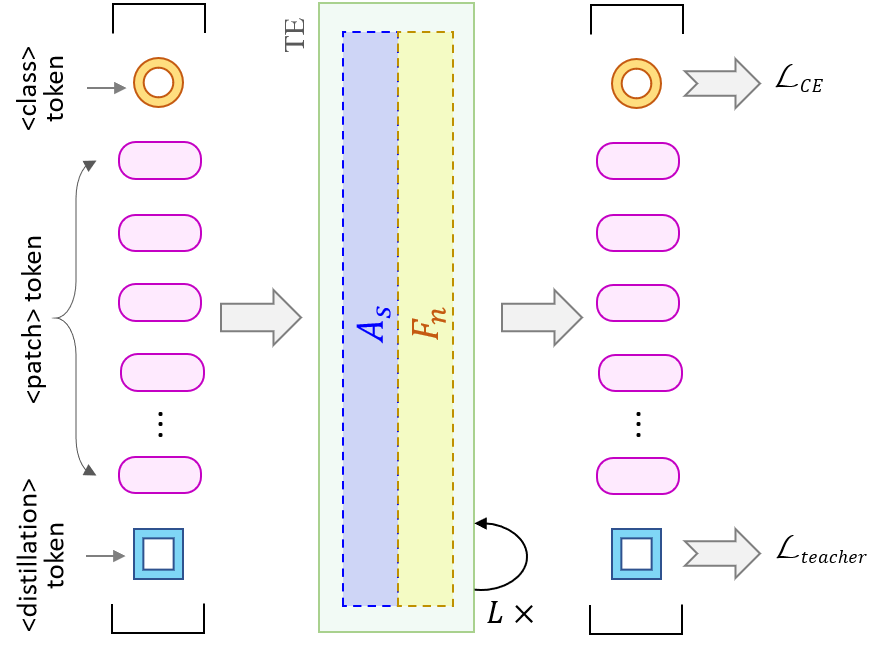}
\caption{The distillation procedure of DeiT}
\label{fig:deit}
\end{figure}

In Fig. \ref{fig:deit}, we present the distillation procedure of DeiT. Here, we add the <distillation> token to the transformer, which interacts with the <class> and <patch> tokens through transformer encoders. 
The transformer encoder used here is similar to the ViT's one, which includes $A_s$ and $F_n$ blocks as shown in Fig. \ref{fig:encoder}.
The objective of the <distillation> token is to reproduce the teacher's predicted label instead of the ground-truth label. The <distillation> and <class> tokens are learned by back-propagation \cite{d2l}.

A linear classifier is used in DeiT instead of the MLP head of ViT \cite{deit,vit} to work efficiently with limited computational resources.

\begin{figure}[!t]
\centering
\includegraphics[width=0.8\linewidth]{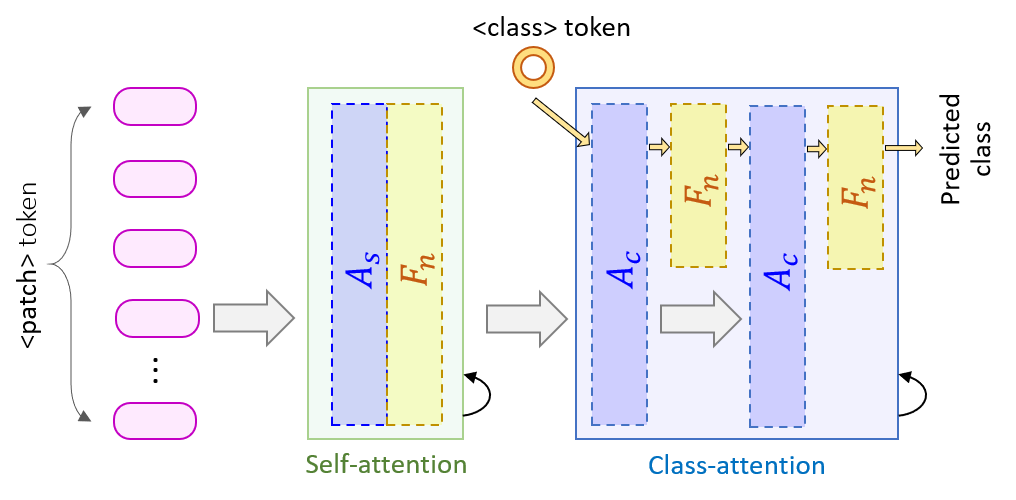}
\caption{Workflow of CaiT}
\label{fig:cait}
\end{figure}

\emph{{\ref{Transformer Models}.3. Class-attention in image Transformers (CaiT)}:} 
\textcolor{black}{CaiT usually performs better than ViT and DeiT with lesser FLOPs and learning parameters \cite{d2l}, when we need to increase the depth of the transformer \cite{cait}. 
CaiT is basically an upgraded version of ViT, which leverages layers with specific class-attention and LayerScale \cite{cait}.} 
In Fig. \ref{fig:cait}, we show the workflow of CaiT.

LayerScale aids CaiT to work at larger depths, where we separately multiply a diagonal matrix $M_\lambda$ on the outputs of $A_s$ and $F_n$ blocks.
\begin{equation}
    \begin{split}
    z_l' = M_\lambda(\lambda_{1}^l,\dots,\lambda_{D}^l) \times MSA(LN(z_{l-1})) + z_{l-1}; \\
    z_l = M_\lambda({\lambda'}_{1}^l,\dots,{\lambda'}_{D}^l) \times MLP(LN(z_l')) + z_l' 
\end{split}
\end{equation}
where, $\lambda_{i}^l$ and ${\lambda'}_{i}^l$ are learning parameters, and other symbols denote the same as the above-mentioned ViT.

In CaiT, the transformer layers dealing with self-attention among patches are separated from class-attention layers that are introduced to dedicatedly extract the content of the patches into a vector, which can be sent to a linear classifier \cite{cait}. 
The <class> token is inserted in the latter stage, so that the initial layers can perform the self-attention among patches devotedly.
In the class-attention stage, we alternatively use multi-head class-attention ($A_c$) \cite{cait} and $F_n$, as shown in Fig. \ref{fig:cait}, and update only the class embedding.

\begin{figure}[!t]
\centering
\includegraphics[width=0.75\linewidth]{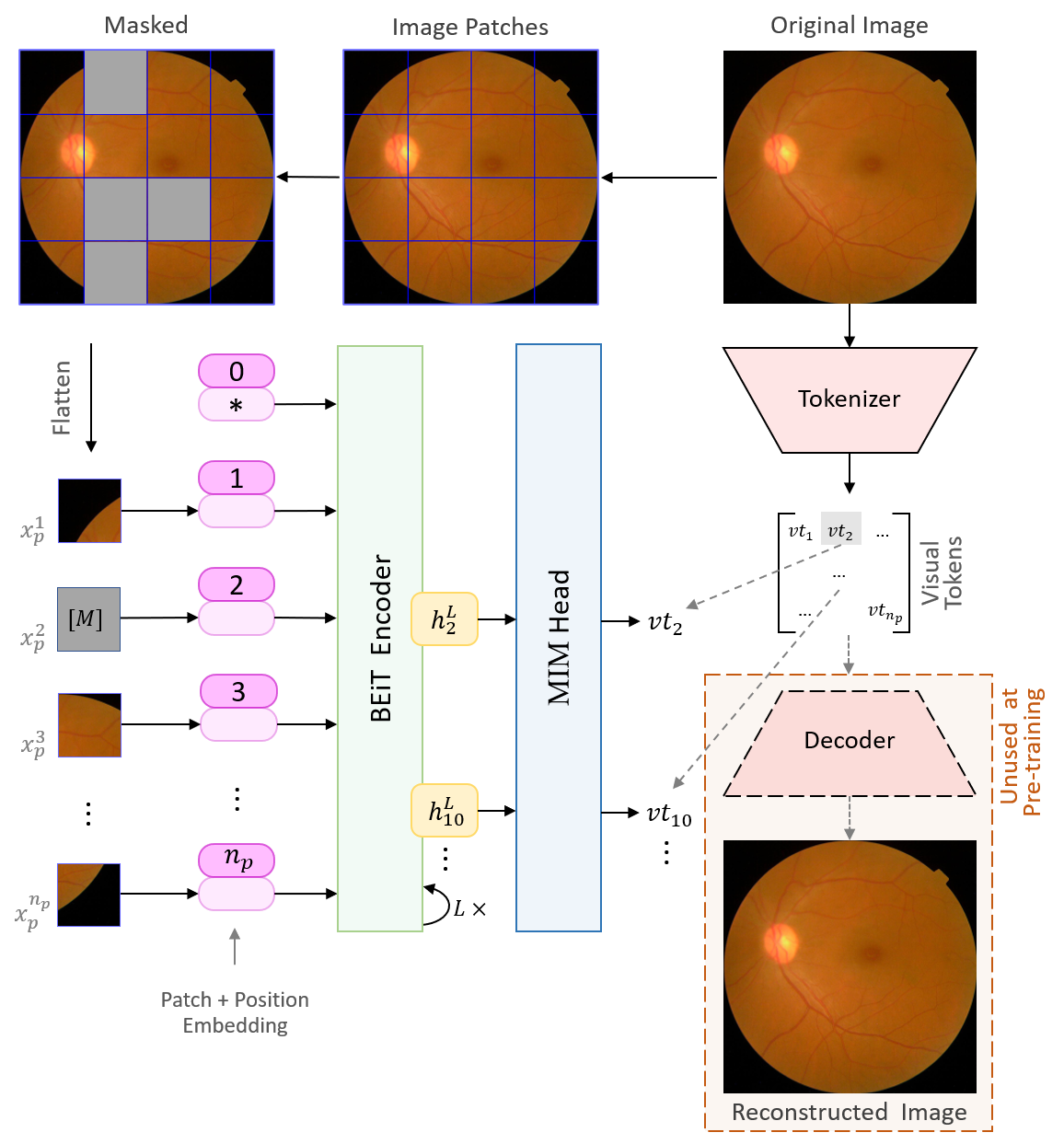}
\caption{Workflow of BEiT pre-training}
\label{fig:beit}
\end{figure}

\emph{{\ref{Transformer Models}.4. Bidirectional Encoder representation for image Transformer (BEiT)}:} 
\textcolor{black}{BEiT is a self-supervised model having its root in the BERT (Bidirectional Encoder Representations from Transformers) \cite{devlin2018bert}, and leverages bidirectional encoding and pre-training \cite{beit}.} 
In Fig. \ref{fig:beit}, we present the workflow of the pre-training of BEiT. 

The input image $I_z$ is split into patches $x_p^i$ and flattened into vectors, similar to the early-mentioned ViT.
In BEiT, a backbone transformer is engaged, for which we use ViT \cite{vit}.
%
On the other hand, $I_z$ is represented as a sequence of visual tokens 
$vt = [vt_1, vt_2, \dots, vt_{n_p}]$
obtained by a discrete VAE (Variational Auto-Encoder) \cite{token}. For visual token learning, we employ a tokenizer 
$\mathcal{T}_\phi(vt~|~x)$ to map image pixels $x$ to tokens $vt$, and decoder $\mathcal{D}_\theta(x~|~vt)$ for reconstructing input image pixels $x$ from $vt$ \cite{beit}.

Here, a MIM (Masked Image Modeling) \cite{beit} task is performed to pre-train the image transformers, where some image patches are randomly masked, and the corresponding visual tokens are then predicted.  
The masked patches are replaced with a learnable embedding $e_{[M]}$.
We feed the corrupted image patches 
$x^\mathcal{M} = \{x_p^i : i \notin \mathcal{M} \} \bigcup \{e_{[M]}: i \in \mathcal{M} \}   $ 
to the transformer encoder. 
Here, $\mathcal{M}$ is the set of indices of masked positions. 

The encoded representation $h_i^L$ is the hidden vector of the last transformer layer $L$ for $i^{th}$ patch.
For each masked position, a softmax classifier $\psi$ is used to predict the respective visual token, i.e., 
$p_\text{MIM}(vt'~|~ x^\mathcal{M}) = \psi(W_M h_i^L + b_M) $; where, $W_M$ and $b_M$ contain learning parameters for linear transformation. 
The pre-training objective of BEiT is to maximize the log-likelihood of the correct token $vt_i$ given $x^\mathcal{M}$, as below:
\[max \sum_{x \in~ \text{DB}_{tr}} \mathds{E}_{\mathcal{M}} \left[\sum_{i \in \mathcal{M}} \text{log}~ p_\text{MIM}\left(vt_i~|~x^\mathcal{M}\right) \right]   \]
where, $\text{DB}_{tr}$ is the training dataset. 
The BEiT pre-training can be perceived as VAE training \cite{beit,token}, where we follow two stages, i.e., 
\emph{stage-1}: minimizing loss for visual token reconstruction,
\emph{stage-2}: modeling masked image, i.e., learning prior $p_\text{MIM}$ by keeping $\mathcal{T}_\phi$ and $\mathcal{D}_\theta$ fixed. It can be written as follows: 
{\small 
\[
\begin{split}
\sum_{\substack{(x_i, x_i^{\mathcal{M}}) \\ \in~ \text{DB}_{tr}}}  \left( 
\underbrace{\mathds{E}_{vt_i \sim \mathcal{T}_\phi(vt|x_i)}  
\left[ \text{log}~ \mathcal{D}_\theta(x_i|vt_i)  \right]}_\text{\emph{stage-1}} 
+ 
\underbrace{\text{log}~ p_\text{MIM}\left(\hat{vt_i}|x_i^\mathcal{M}\right)}_\text{\emph{stage-2}}
\right)  
\end{split}
\]
}
where, $\hat{vt_i} = \text{argmax}_{vt} ~\mathcal{T}_\phi(vt~|~x_i)  $.


\subsubsection{Ensembled Transformers}
\label{Ensembled Transformers}
\begin{figure}[!h]
\centering
\includegraphics[width=0.75\linewidth]{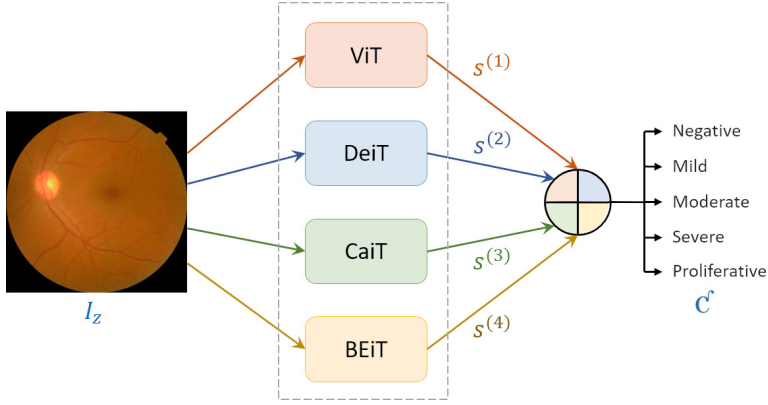}
\caption{\textcolor{black}{Ensembled transformer network.}}
\label{fig:ensemble}
\end{figure}

\textcolor{black}{The abovementioned four image transformers, 
i.e., ViT \cite{vit},
DeiT \cite{deit}, 
CaiT \cite{cait}, and 
BEiT \cite{beit} 
are pre-trained on the training set DB$_{tr}$. 
We now ensemble the transformers for predicting the severity stages from fundus images of the test set DB$_t$, 
since ensembling multiple learning algorithms can achieve better performance than the constituent algorithms alone \cite{ensemble}. The pictorial representation of ensembled transformer network is presented in Fig. \ref{fig:ensemble}.}

For an image sample from DB$_t$, we obtain the softmax probability distribution $s^{(j)} : \{ P_1^j, P_2^j, \dots, P_{n_c}^j \}$ over $j^{th}$ transformer \cite{d2l}, for $j=1,2, \dots, n_T$;
where, $n_c$ is the total number of classes (severity stages), and $n_T$ is count of the employed image transformers. 
Here,
$\sum_{i=1}^{n_c}P_i^j=1$, $n_c=5$ (refer to subsection \ref{subsec:problemFormulation}), and 
$n_T=4$ since we use four separately trained distinct image transformers, as mentioned earlier.

We obtain the severity stages/ class\_labels 
\texthtc$|_{wm}$ and \texthtc$|_{mv}$ separately using two combination methods \emph{weighted mean} and \emph{majority voting} \cite{ensemble}, respectively. 
\begin{equation}
    \text{\texthtc}|_{wm} = \text{argmax}_i ~ P_i^\mu ~;~ 
    P_i^\mu = \frac{\sum_{j=1}^{n_T}{\alpha_j P_i^j}}{\sum_{j=1}^{n_T}\alpha_j} ~;~
    \text{for } i=1,2,\dots,n_c
\label{eq:weight}
\end{equation}
In this task, we choose $\sum_{j=1}^{n_T}\alpha_j=1$. 
\begin{equation}
\begin{split}
\text{\texthtc}|_{mv} = \text{mode}\left(\text{argmax}_i(P_i^1), \text{argmax}_i(P_i^2), \dots, \text{argmax}_i(P_i^{n_T})\right)\\
= \text{mode}\left(\text{argmax}_i(s^{(1)}), \text{argmax}_i(s^{(2)}), \dots, \text{argmax}_i(s^{(n_T)})\right) ; \\
\text{for } i=1,2,\dots,n_c
\end{split}
\end{equation}

\textcolor{black}{In this task, we use cross-entropy as the loss function \cite{goodfellow2016deep} in the employed image transformers. The AdamW optimizer is used here due to its weight decay regularization effect for tackling overfitting \cite{adamw}.} 
The training details with hyper-parameter tuning are mentioned in Section \ref{subsec:result}.

\section{Experiments and Discussions}
\label{experimental_result}
In this section, we present the employed database, followed by experimental results with discussions.

\subsection{Database Employed}
\label{subsec:database}
\begin{figure}[!t]
\centering
\includegraphics[width=0.75\linewidth]{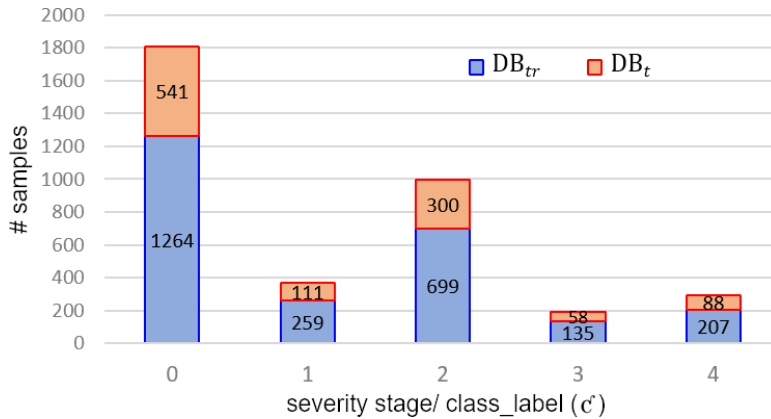}
\caption{Count of samples in APTOS-2019 \cite{aptos2019}.}
\label{fig:aptos_count}
\end{figure}

For our computational experiments, we used the publicly available training samples of Kaggle APTOS (\emph{Asia Pacific Tele-Ophthalmology Society}) 2019 Blindness Detection dataset \cite{aptos2019}, 
i.e., APTOS-2019. 
This database (DB) contains fundus image samples of five severity stages of DR, i.e., negative, mild, moderate, severe, and proliferative. 
Fig. \ref{fig:samples} shows some sample images from this dataset.
\textcolor{black}{In DB, a total of 3662 fundus images are available, which we divide into training (DB$_{tr}$) and testing (DB$_t$) datasets with a ratio of $7:3$, 
ensuring that both DB$_{tr}$ and DB$_t$ include samples of each stratum (DR severity class) in the same ratio.} 
As a matter of fact, DB$_{tr}$ and DB$_t$ sets are disjoint. For validating our model, 10\% data of DB$_{tr}$ are kept as the validation set DB$_v$. 
\textcolor{black}{The sample counts of different severity stages/ class\_labels (\texthtc) for DB$_{tr}$ and DB$_t$ are shown in Fig. \ref{fig:aptos_count} individually. 
Here, 49.3\% of the samples belong to the \emph{negative} DR category (\texthtc = 0). 
Among the positive classes, the \emph{moderate} stage (\texthtc = 2) constitutes the largest portion, accounting for 27.28\% of the total sample count, while the \emph{severe} stage (\texthtc = 3) represents the smallest, with only 5.27\% of the total samples. 
Fig. \ref{fig:aptos_count} depicts the data imbalance in DB due to varying sample counts across different severity stages. To address this imbalance, we employed data augmentation during model training, as mentioned in subsection \ref{Preprocessing}. This data augmentation also effectively mitigated the risk of overfitting \cite{d2l}.}

\subsection{Experimental Results}
\label{subsec:result}
This section discusses the performed experiments, analyzes the model outcome, and compares them with major state-of-the-art methods. We begin with discussing the experimental settings. 

\subsubsection{Experiment Settings} 
We performed the experiments on the TensorFlow-2 framework having Python 3.7.13 over a machine with the following configurations: Intel(R) Xeon(R) CPU @ 2.00GHz with 52 GB RAM and Tesla T4 16 GB GPU. 
All the results shown here were obtained from DB$_t$. 

\textcolor{black}{The hyper-parameters of the framework were tuned and fixed during training with respect to the performance over samples of DB$_v$. 
For all the image transformers used here (i.e., ViT, DeiT, CaiT, and BEiT), we empirically set the following hyper-parameters: 
transformer\_layers ($L$) = 12, 
embedding\_dimension ($D$) = 384, 
num\_heads ($h$) = 6. 
The following hyper-parameters were selected for AdamW \cite{adamw}: 
initial\_learning\_rate = $10^{-3}$; 
exponential decay rates for 1$^{st}$ and 2$^{nd}$ moment estimates, i.e., $\beta1=0.9$, $\beta2=0.999$;
zero-denominator removal parameter ($\upvarepsilon$) = $10^{-8}$; and 
weight\_decay = $10^{-3}/4$.
For model training, the mini-batch size was fixed to 32.}

\subsubsection{Model Performance} 
\label{subsubsec:model_performance}
In Table \ref{tab:ensemble}, we present the performance of our ensembled image transformer ($EiT$) using the combination schemes weighted mean ($wm$) and majority voting ($mv$), where we obtain
94.63\% and 91.26\% accuracy from $EiT_{wm}$ and $EiT_{mv}$, respectively. 
We also ensembled multiple combinations of our employed transformers, and present their performances in this table. Here, the $wm$ scheme performed better than $mv$. As evident from this table, ensembling various types of transformers improved the performance. 
Among single transformers (for $n_T=1$), CaiT performed the best.
For $n_T=2$ and $n_T=3$, 
{\textquotedblleft{BEiT + CaiT}\textquotedblright} and 
{\textquotedblleft{DeiT + BEiT + CaiT}\textquotedblright} performed better than other respective combinations. 
Overall, $EiT_{wm}$ attained the best accuracy here.

\begin{table}[!h]
\footnotesize
\centering
\caption{Performance over various ensembling of transformers}
    \begin{tabular}{c|c|c|c}
\hline 
\multirow{3}{*}{$n_T$} & \multirow{3}{*}{Ensembled Transformers} & \multicolumn{2}{c}{Accuracy (\%)} \\ \cline{3-4}
 &    & {Weighted} & {Majority} \\ 
&    & {mean} & {voting} \\ \hline \hline
\multirow{4}{*}{1} & ViT & \multicolumn{2}{c}{82.21}  \\ \cline{2-4}
& DeiT & \multicolumn{2}{c}{85.65}  \\ \cline{2-4}
& BEiT & \multicolumn{2}{c}{86.74} \\ \cline{2-4}
& CaiT & \multicolumn{2}{c}{86.91} \\ \hline
\multirow{6}{*}{2} & ViT + DeiT & 87.03 & 86.55 \\ \cline{2-4}
& ViT + BEiT & 87.48 & 87.03 \\ \cline{2-4}
& ViT + CaiT & 87.77 & 87.21 \\ \cline{2-4}
& DeiT + BEiT & 88.18 & 87.69 \\ \cline{2-4}
& DeiT + CaiT & 88.86 & 87.93 \\ \cline{2-4}
& BEiT + CaiT & 89.28 & 88.12 \\ \hline
\multirow{4}{*}{3} & ViT + DeiT + BEiT & 90.53 & 88.87 \\ \cline{2-4}
& ViT + DeiT + CaiT & 91.39 & 89.56 \\ \cline{2-4}
& ViT + BEiT + CaiT & 92.14 & 90.28 \\ \cline{2-4}
& DeiT + BEiT + CaiT & 93.46 & 90.91 \\ \hline
\multirow{2}{*}{4} & ViT + DeiT + BEiT + CaiT & \multirow{2}{*}{94.63} & \multirow{2}{*}{91.26}\\
& ( $EiT$ )   & &
\\ \hline
    \end{tabular}
    \label{tab:ensemble}
\end{table}

In Fig. \ref{fig:gradcam} of \ref{app:gradcam}, we present the coarse localization maps generated by Grad-CAM \cite{gradcam} from the employed individual image transformers to highlight the crucial regions for understanding the severity stages.

\emph{{\ref{subsubsec:model_performance}.1. Various Evaluation Metrics}:}
\textcolor{black}{Besides the accuracy, in Table \ref{tab:metric}, we present the performance of $EiT$ with respect to some other evaluation metrics, e.g., kappa score, precision, recall (or sensitivity), F$_1$ score, specificity, balanced accuracy \cite{metrics}. 
Here, Cohen’s quadratic weighted kappa measures the agreement between human-assigned scores (i.e., DR severity stages) and the $EiT$-predicted scores. 
Precision analyzes the true positive samples among the total positive predictions. 
Recall or sensitivity finds the true positive rate. 
Similarly, specificity computes the true negative rate. 
F$_1$ score is the harmonic mean of precision and recall. 
Since the employed DB is imbalanced, we also compute the balanced accuracy, which is the arithmetic mean of sensitivity and specificity.  
In this table, we can see that for both $EiT_{wm}$ and $EiT_{mv}$, the kappa scores are greater than 0.81, which comprehends the 
{\textquotedblleft{almost perfect agreement}\textquotedblright} between the human rater and $EiT$ \cite{metrics}. Here, \emph{macro} means the arithmetic mean of all per class precision/ recall/ F$_1$ score.} 

\begin{table}[!ht]
\footnotesize
    \centering
\caption{Performance of $EiT$ over various evaluation metrics}
    \begin{tabular}{c|c|c}
    \hline 
\multirow{2}{*}{Metric} & {Weighted mean} & {Majority voting} \\
   &  ($EiT_{wm}$)   & ($EiT_{mv}$) \\ \hline \hline                          
Accuracy (\%)           & 94.63 & 91.26\\ \hline
Kappa score             & 0.92 & 0.87\\ \hline
Macro Precision (\%)     & 90.55 & 84.65\\ \hline
Macro Recall (\%)       & 92.88 & 88.81\\ \hline
Macro F$_1$-score (\%)  & 91.67 & 86.55\\ \hline
Macro Specificity (\%)  & 98.62 & 97.74\\ \hline
Balanced Accuracy (\%)   & 95.75 & 93.27\\ \hline
    \end{tabular}
    \label{tab:metric}
\end{table}

\emph{{\ref{subsubsec:model_performance}.2. Individual Class Performance}:}
Table \ref{tab:per_class} presents the individual performance of $EiT_{wm}$ and $EiT_{mv}$ for detecting every severity stage of DR. 
\textcolor{black}{From this table, we can see our models exhibited the highest precision and recall in identifying \emph{negative} DR (\texthtc = 0) category, while the lowest performance was for \emph{severe} DR class (\texthtc = 3). This reduced performance in the severe class is attributed to the scarcity of available samples within this category (refer to Section \ref{subsec:database}).} 

\begin{table}[!h]
\centering
\scriptsize
\caption{Performance of $EiT$ on every DR severity stage}
    \begin{tabular}{c|c|c|c|c|c|c}
\cline{2-7}
& class\_label (\texthtc) & 0 & 1 & 2 & 3 & 4 \\ \hline \hline 
\multirow{4}{*}{\rotatebox{90}{$EiT_{wm}$}} & Precision (\%)  & \textbf{98.48} & 86.67 & \emph{95.00} & \underline{83.61} & 89.01  \\ \cline{2-7} 
& Recall (\%)     & \textbf{95.75} & 93.69 & \emph{95.00} & \underline{87.93} & 92.05  \\ \cline{2-7} 
& F$_1$-score (\%) & \textbf{97.09} & 90.04 & \emph{95.00} & \underline{85.71} & 90.50  \\ \cline{2-7} 
& Specificity (\%) & 98.56 & 98.38 & \underline{98.12} & \textbf{99.04} & \emph{99.01}  \\ \hline
\multirow{4}{*}{\rotatebox{90}{$EiT_{mv}$}} & Precision (\%)  & \textbf{96.74} & 79.67 & \emph{94.14} & \underline{70.59} & 82.11 \\ \cline{2-7}
& Recall (\%) & \textbf{93.35} & 88.29 & \emph{91.00} & \underline{82.76} & 88.64 \\ \cline{2-7}
& F$_1$-score (\%)  & \textbf{95.01} & 83.76 & \emph{92.54} & \underline{76.19} & 85.25 \\ \cline{2-7}
& Specificity (\%) & \underline{96.95} & 97.47 & 97.87 & \emph{98.08} & \textbf{98.32} \\ \hline 
\multicolumn{7}{r}{\tiny{In each row, the best result is marked \textbf{bold}, second-best is \emph{italic}, and lowest is \underline{underlined}.}}
    \end{tabular}
    \label{tab:per_class}
\end{table}

\subsubsection{\textcolor{black}{Comparison}} 


\textcolor{black}{In Table \ref{tab:comparison}, we present a comparative analysis with some major contemporary CNN-based architectures, such as 
ResNet50 \cite{resnet}, 
InceptionV3 \cite{inceptionNet}, 
MobileNetV2 \cite{mobilenetv2}, 
Xception \cite{xception} and its modified version (Kassani et al. \cite{kassani2019}), 
EfficientNet \cite{efficientnet}, 
SE-ResNeXt50 \cite{squeeze}, and 
ensembled CNN (Tymchenko et al. \cite{Tymchenko}).
CNN-based classifiers often lack the ability to focus on specific regions/ features within fundus images while disregarding remaining portions, potentially leading to a loss of spatial relationships in learned features for DR severity detection. 
Prior attempts on incorporating attention mechanisms can be seen, e.g., Farag et al. \cite{farag2022automatic} used of DenseNet169 with attention, TAN (Texture Attention Network) \cite{alahmadi2022texture} aimed to recalibrate texture and content features. However, these attention-based models did not consistently demonstrate superior performance. 
In response to these limitations, recent advancements introduced transformer-based models featuring multi-head self-attention. For instance, CoT-XNet \cite{cotxnet} integrated contextual transformer with Xception, and SSiT \cite{ssit} employed self-supervised image transformers guided by saliency maps, which have shown promise in utilizing transformer networks for DR severity detection from fundus images. 
Our weighted mean-based ensembled transformer network $EiT_{wm}$ outperformed the major state-of-the-art methods with respect to accuracy, balanced accuracy, sensitivity, and specificity. 
Our $EiT_{mv}$ also performed quite well in terms of balanced accuracy.}

\begin{table}[!t]
    \centering
\scriptsize
\caption{\textcolor{black}{Comparative study}}
    \begin{tabular}{l|c|c|c|c}
\hline
\multirow{2}{*}{Method} & Accuracy & Sensitivity & Specificity & Balanced \\
                &   (\%)      &    (\%)     &    (\%)     & Accuracy (\%)
\\ \hline \hline
ResNet50 \cite{resnet}   & 74.64 & 56.52 & 85.71 & 71.12 \\ \hline
InceptionV3 \cite{inceptionNet}         & 78.72 & 63.64 & 85.37 & 74.51 \\ \hline
MobileNetV2 \cite{mobilenetv2}   & 79.01 & 76.47 & 84.62 & 80.55 \\ \hline
Xception \cite{xception}    & 79.59 & 82.35 & 86.32 & 84.34 \\ \hline
Kassani et al. \cite{kassani2019}   & 83.09 & 88.24 & 87.00 & 87.62 \\ \hline
EfficientNet-B4 \cite{efficientnet}     & 90.30 & 81.20 & 97.60 & 89.40 \\ \hline
EfficientNet-B5 \cite{efficientnet}     & 90.70 & 80.70 & 97.70 & 89.20 \\ \hline
SE-ResNeXt50 \cite{squeeze}  & 92.40 & 87.10 & 98.20 & 92.65 \\ \hline
Tymchenko et al. \cite{Tymchenko} & 92.90 & 86.00 & \underline{98.30} & 92.15 \\ \hline
Farag et al. \cite{farag2022automatic}   & 82.00 & - & - & - \\ \hline
TAN \cite{alahmadi2022texture} & 85.10 & 90.30 & 92.00 & - \\ \hline
CoT-XNet \cite{cotxnet}  & 84.18 & - & 95.74 & - \\ \hline
SSiT \cite{ssit}        & \underline{92.97} & - & - & - \\ \hline
$EiT_{mv}$ [ours]   & 91.26 & \underline{88.81} & 97.74 & \underline{93.28} \\ \hline
$EiT_{wm}$ [ours]   & \textbf{94.63} & \textbf{92.88} & \textbf{98.62} & \textbf{95.75} \\ \hline
\multicolumn{5}{r}{In each column, the best result is marked \textbf{bold}, and the second-best is \underline{underlined}.}
    \end{tabular}
    \label{tab:comparison}
\end{table}

\subsubsection{\textcolor{black}{Impact of Hyper-parameters}}
\label{subsubsec:hyperparam}
We tuned the hyper-parameters and observed their impact on the experiment.

\begin{figure}[!t]
\centering
\includegraphics[width=0.8\linewidth]{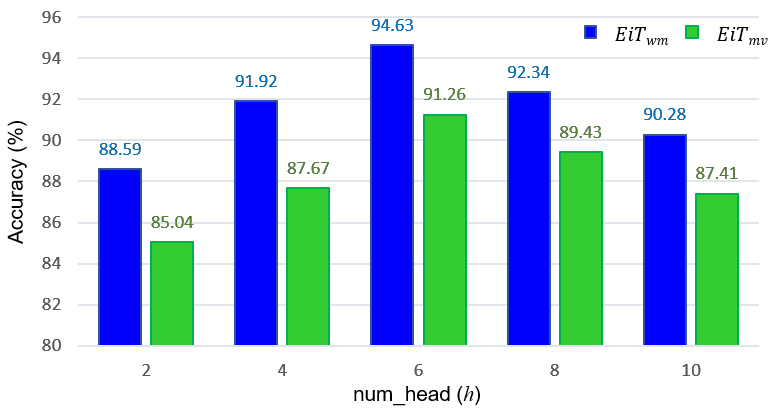}
\caption{Impact of number of heads ($h$) in MSA on model performance}
\label{fig:head}
\end{figure}

\emph{{a) MSA Head Count}:}
We analyzed the performance impact of the number of heads ($h$) of MSA (Multi-head Self-Attention) in the transformer encoder and present in Fig. \ref{fig:head}.
As evident from this figure, the performance (accuracy) of both $EiT_{mv}$ and $EiT_{wm}$ increased with the increment of $h$ till $h=6$, and started decreasing thereafter. 

\emph{{b) Weights $\alpha_j$ of $EiT_{wm}$}:}
We tuned the weights $\alpha_j$ (refer to Eqn. \ref{eq:weight}) to see its impact on the performance of $EiT_{wm}$. 
\textcolor{black}{Here, $\alpha_j$'s were tuned by grid-search technique \cite{hpTuning}.}
We obtained the best accuracy of 94.63\% from $EiT_{wm}$ for $\alpha_1=\alpha_2=0.1$, and $\alpha_3=\alpha_4=0.4$. 
The performance of $EiT_{wm}$ during tuning of $\alpha_j$'s is shown in Table \ref{tab:weights1}. 
In Table \ref{tab:weights2} of \ref{app:tuned_weights}, 
we also present the tuned $\alpha_j$'s that aided in obtaining the best performing ensembled transformers of Table \ref{tab:ensemble}.

\begin{table}[!t]
\centering
\scriptsize
\caption{Performance of $EiT_{wm}$ by tuning weights $\alpha_j$}
    \begin{tabular}{c|c|c|c|c}
\hline
$\alpha_1$ & $\alpha_2$ & $\alpha_3$ & $\alpha_4$ & Accuracy (\%) \\ \hline \hline
0.25 & 0.25 & 0.25 & 0.25 & 89.53\\ \hline
0.85 & 0.05 & 0.05 & 0.05 & 82.29\\ \hline
0.05 & 0.85 & 0.05 & 0.05 & 85.78\\ \hline
0.05 & 0.05 & 0.85 & 0.05 & 86.92\\ \hline
0.05 & 0.05 & 0.05 & 0.85 & 87.05\\ \hline
0.7 & 0.1 & 0.1 & 0.1 & 82.35\\ \hline
0.1 & 0.7 & 0.1 & 0.1 & 85.91\\ \hline
0.1 & 0.1 & 0.7 & 0.1 & 87.04\\ \hline
0.1 & 0.1 & 0.1 & 0.7 & 87.20\\ \hline
0.5 & 0.167 & 0.167 & 0.166 & 82.88\\ \hline
0.166 & 0.5 & 0.167 & 0.167 & 86.35\\ \hline
0.167 & 0.166 & 0.5 & 0.167 & 87.62\\ \hline
0.167 & 0.167 & 0.166 & 0.5 & 87.74\\ \hline
0.3 & 0.3 & 0.2 & 0.2 & 88.16\\ \hline
0.3 & 0.2 & 0.3 & 0.2 & 89.58\\ \hline
0.3 & 0.2 & 0.2 & 0.3 & 90.27\\ \hline
0.2 & 0.3 & 0.3 & 0.2 & 90.85\\ \hline
0.2 & 0.3 & 0.2 & 0.3 & 91.67\\ \hline
0.2 & 0.2 & 0.3 & 0.3 & 92.72\\ \hline
0.4 & 0.4 & 0.1 & 0.1 & 91.18\\ \hline
0.4 & 0.1 & 0.4 & 0.1 & 91.49\\ \hline
0.4 & 0.1 & 0.1 & 0.4 & 92.15\\ \hline
0.1 & 0.4 & 0.4 & 0.1 & 92.84\\ \hline
0.1 & 0.4 & 0.1 & 0.4 & 93.47\\ \hline
0.1 & 0.1 & 0.4 & 0.4 & \textbf{94.63}\\ \hline
    \end{tabular}
    \label{tab:weights1}
\end{table}

\subsubsection{\textcolor{black}{Ablation Study}}
\label{subsubsec:ablation}

We here present the conducted ablation study, where we systematically ablate individual transformers to assess their impact. 
Our $EiT$ is actually an ensembling of four different image transformers, i.e., ViT, DeiT, CaiT, and BEiT. 
We ablated each transformer and observed performance degradation than $EiT$. 
For example, considering the weighted mean scheme, when we ablated CaiT from $EiT$, the accuracy dropped by 4.1\%. Similarly, ablating BEiT and CaiT deteriorated the accuracy by 7.6\%. For our task, the best individual transformer (CaiT) attained 7.72\% lower accuracy than $EiT_{wm}$. More examples can be observed in Table \ref{tab:ensemble}.

\subsubsection{\textcolor{black}{Pre-training with Other Datasets}} 
\label{subsubsec:pre_training}
We checked the performance of our $EiT$ model by pre-training with some other dataset. 
We took 1200 images of MESSIDOR \cite{Messidor2014} with adjudicated grades by \cite{Google2018grader} (say, DB$_M$). 
From IDRiD \cite{idrid2018}, we also used {\textquotedblleft{Disease Grading}\textquotedblright} dataset containing  516 images (say, DB$_I$).
Here, we made four training set setups from DB$_M$, by taking 25\%, 50\%, 75\%, and 100\% of samples of DB$_M$. 
Similarly, four training setups were generated from DB$_I$. 
As mentioned in subsection \ref{subsec:database}, we divided APTOS-2019 database (DB) in training (DB$_{tr}$) and test (DB$_t$) sets with a ratio of $7:3$.
In Table \ref{tab:pretraining}, we present the performance of $EiT$ on DB$_t$, while pre-training with DB$_M$ and DB$_I$, and training with DB$_{tr}$. 
It can be observed that the performance of $EiT$ improved slightly when pre-trained with more data from other datasets.

\begin{table}[!htb]
\centering
\scriptsize
\caption{Accuracy (\%) of $EiT$ with pre-training}
    \begin{tabular}{c|c|  c|c|c|c}
\cline{2-6}
& Pre-training data & 25\% & 50\% & 75\% & 100\% \\ \hline \hline 
\multirow{4}{*}{\rotatebox{90}{$EiT_{wm}$}} & DB$_M$    &  94.71 & 94.78 & 94.83 & 94.88  \\ \cline{2-6} 
& DB$_I$    &  94.65 & 94.67 & 94.7 & 94.79  \\ \cline{2-6} 
& DB$_M$ + DB$_I$  &  94.73 & 94.85 & 94.98 & 95.13 \\ \cline{2-6}
& \emph{N.A.} & \multicolumn{4}{c}{94.63}  \\ \hline
\multirow{4}{*}{\rotatebox{90}{$EiT_{mv}$}} & DB$_M$    &  91.35 & 91.48 & 91.56 & 91.61 \\ \cline{2-6} 
& DB$_I$    & 91.27 & 91.32 & 91.34 & 91.35
  \\ \cline{2-6} 
& DB$_M$ + DB$_I$  &  91.42 & 91.6 & 91.68 & 91.75
 \\ \cline{2-6}
& \emph{N.A.} & \multicolumn{4}{c}{91.26}  \\ \hline
\multicolumn{6}{r}{\emph{N.A.}: without pre-training data}
    \end{tabular}
    \label{tab:pretraining}
\end{table}

\section{Conclusion}\label{conc}
In this paper, we tackle the problem of automated severity stage detection of DR from fundus images.
For this purpose, we propose two ensembled image transformers, $EiT_{wm}$ and $EiT_{mv}$, by using weighted mean and majority voting combination schemes, respectively. 
We here adopt four transformer models, 
i.e., ViT, DeiT, CaiT, and BEiT. 
For experimentation, we employed the publicly available APTOS-2019 blindness detection dataset, on which $EiT_{wm}$ and $EiT_{mv}$ attained accuracies of 94.63\% and 91.26\%, respectively. Although the employed dataset was imbalanced, our models performed quite well. 
Our $EiT_{wm}$ outperformed the major state-of-the-art techniques. 
We also performed an ablation study and observed the importance of the ensembling over the individual transformers. 

In the future, we will endeavor to improve the model performance with some imbalanced learning techniques. 
Currently, our model does not perform any lesion segmentation, which we will also attempt in order to explore some implicit characteristics of fundus images due to DR.

\appendix 

\section{Qualitative Visualization}
\label{app:gradcam}

\begin{figure}[!htb]
\small
\centering
\footnotesize
\begin{adjustbox}{width=\textwidth}
\begin{tabular}{c|c|c|c|c|c}
\hline 
&  && &&  \\[\dimexpr-\normalbaselineskip+1.5pt]
\rotatebox{90}{Fundus}    &{\includegraphics[width=.0875\textwidth, height=.09\textwidth]{images/Ground_truth/0.jpg}} &
    {\includegraphics[width=.0875\textwidth, height=.09\textwidth]{images/Ground_truth/1.jpg}} &
    {\includegraphics[width=.0875\textwidth, height=.09\textwidth]{images/Ground_truth/2.jpg}} &
    \includegraphics[width=.0875\textwidth, height=.09\textwidth]{images/Ground_truth/3.jpg} &
    \includegraphics[width=.0875\textwidth, height=.09\textwidth]{images/Ground_truth/4.jpg}\\ \hline
 & && && \\[\dimexpr-\normalbaselineskip+1.5pt]
\rotatebox{90}{~~ViT}   &{\includegraphics[width=.0875\textwidth, height=.09\textwidth]{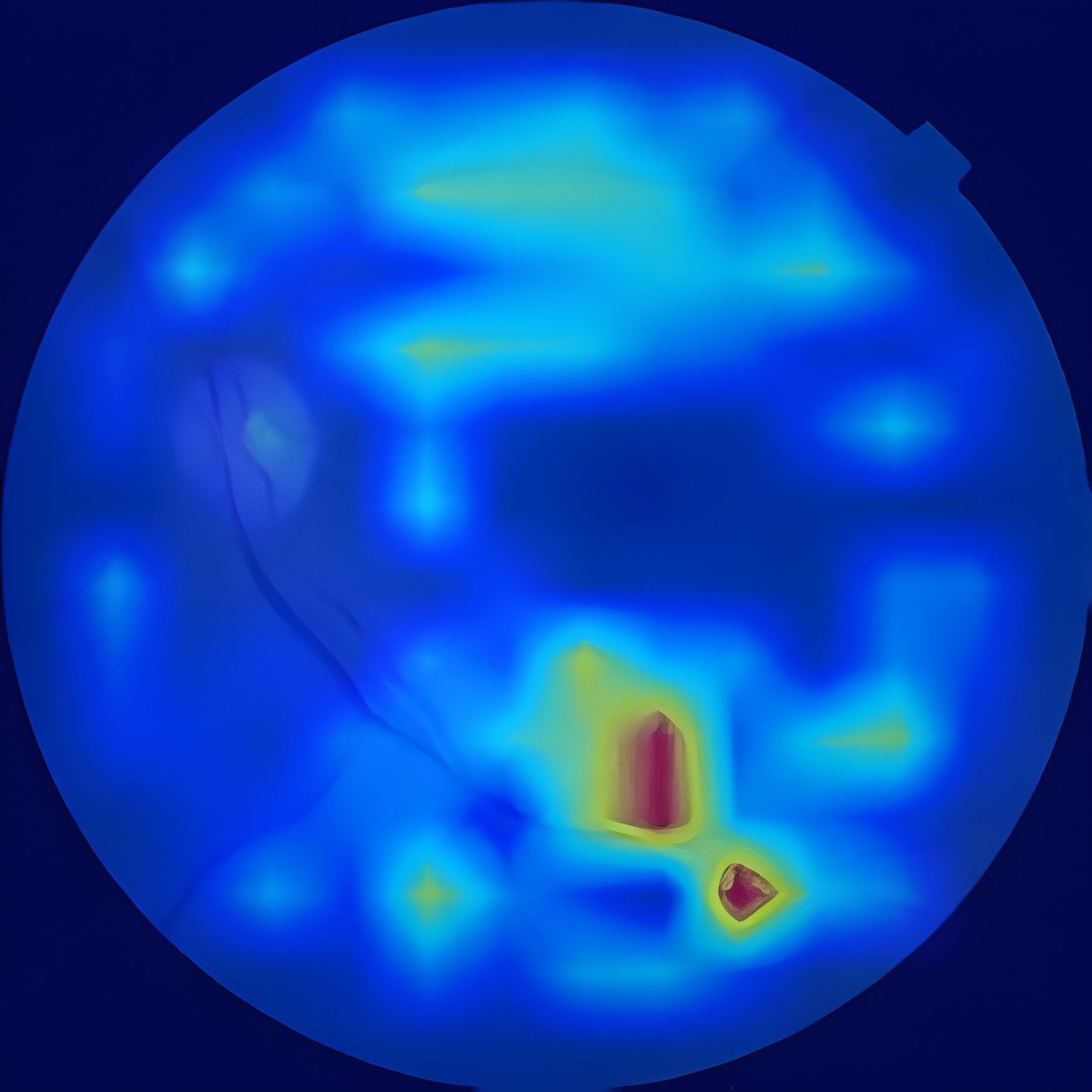}} &
    {\includegraphics[width=.0875\textwidth, height=.09\textwidth]{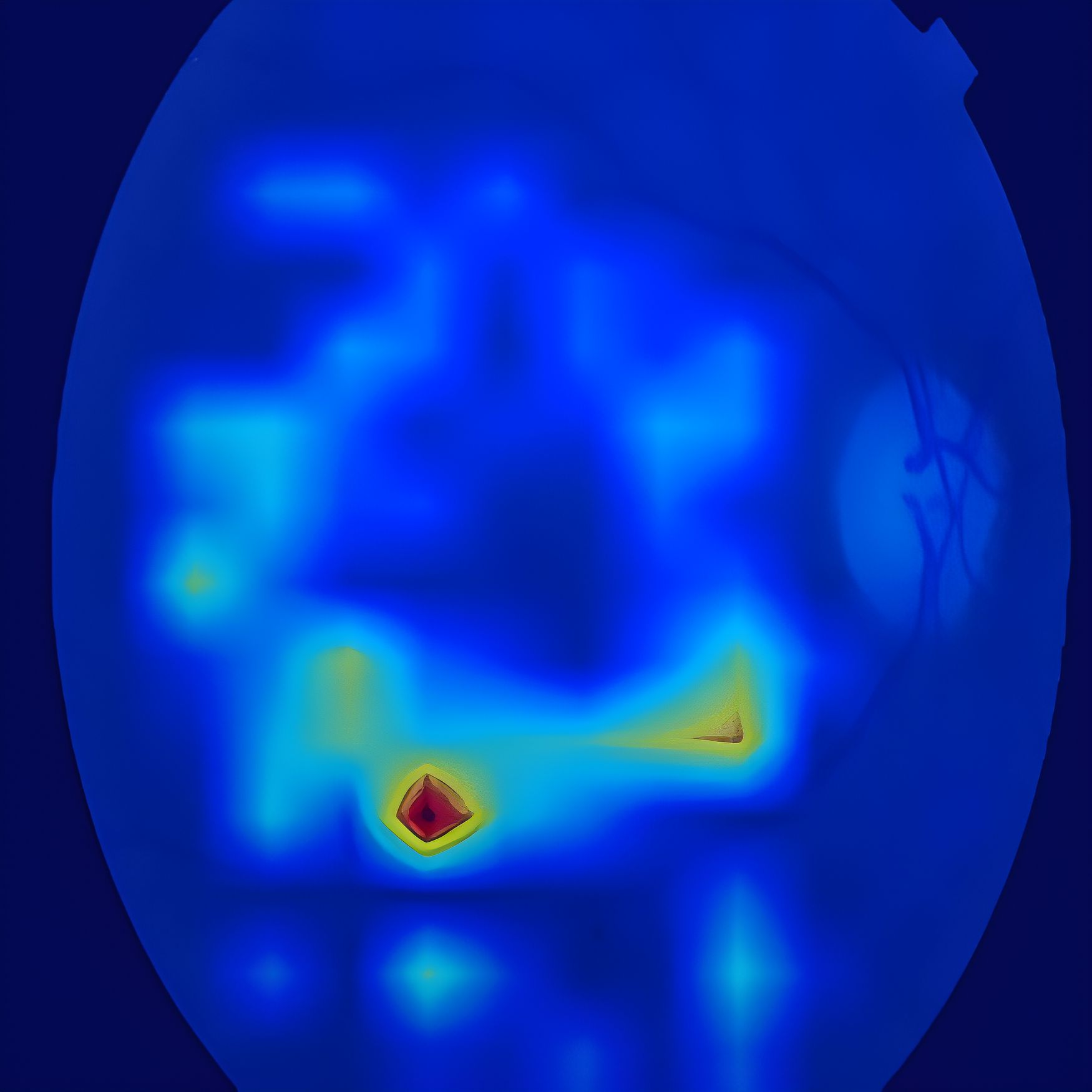}} &
    {\includegraphics[width=.0875\textwidth, height=.09\textwidth]{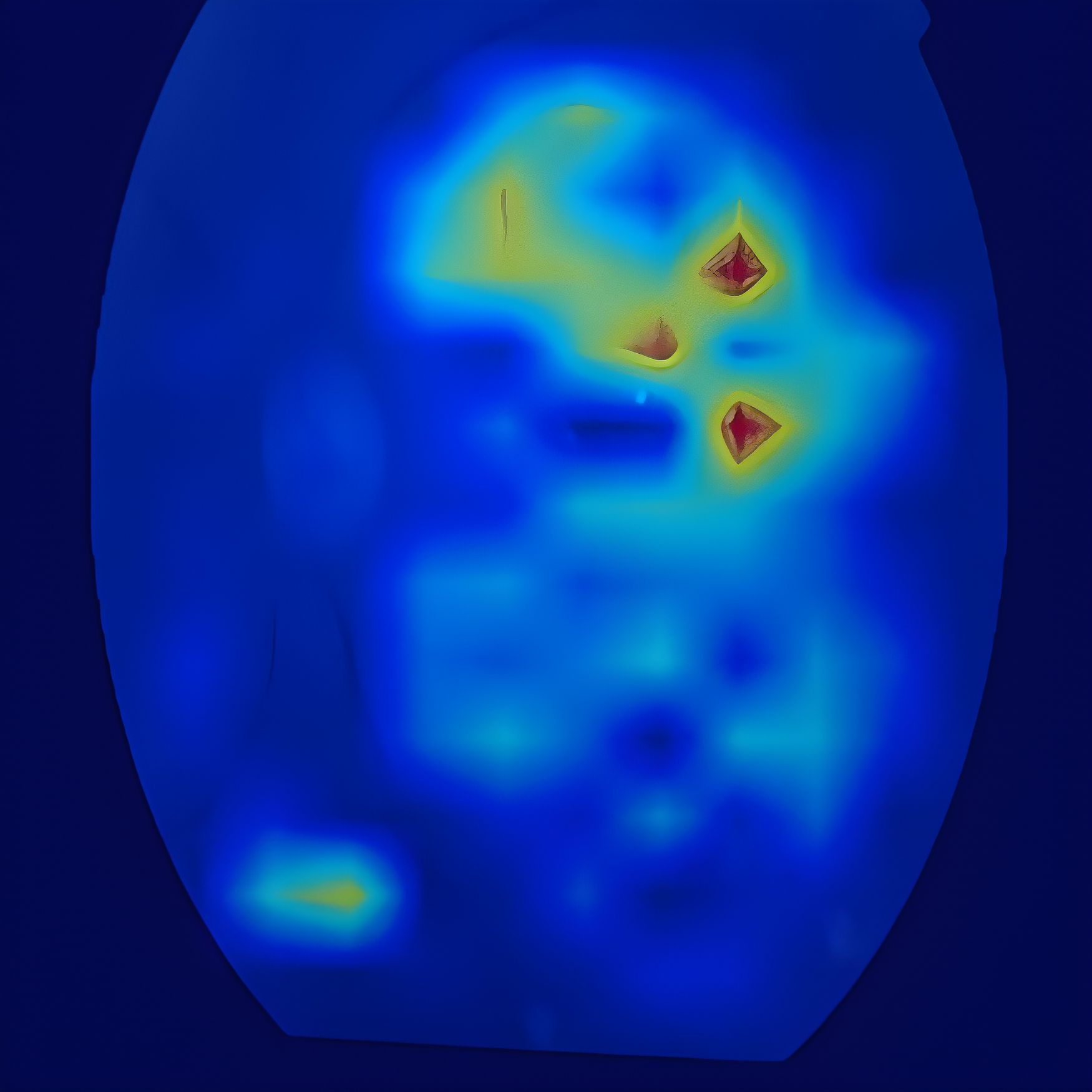}} &
    \includegraphics[width=.0875\textwidth, height=.09\textwidth]{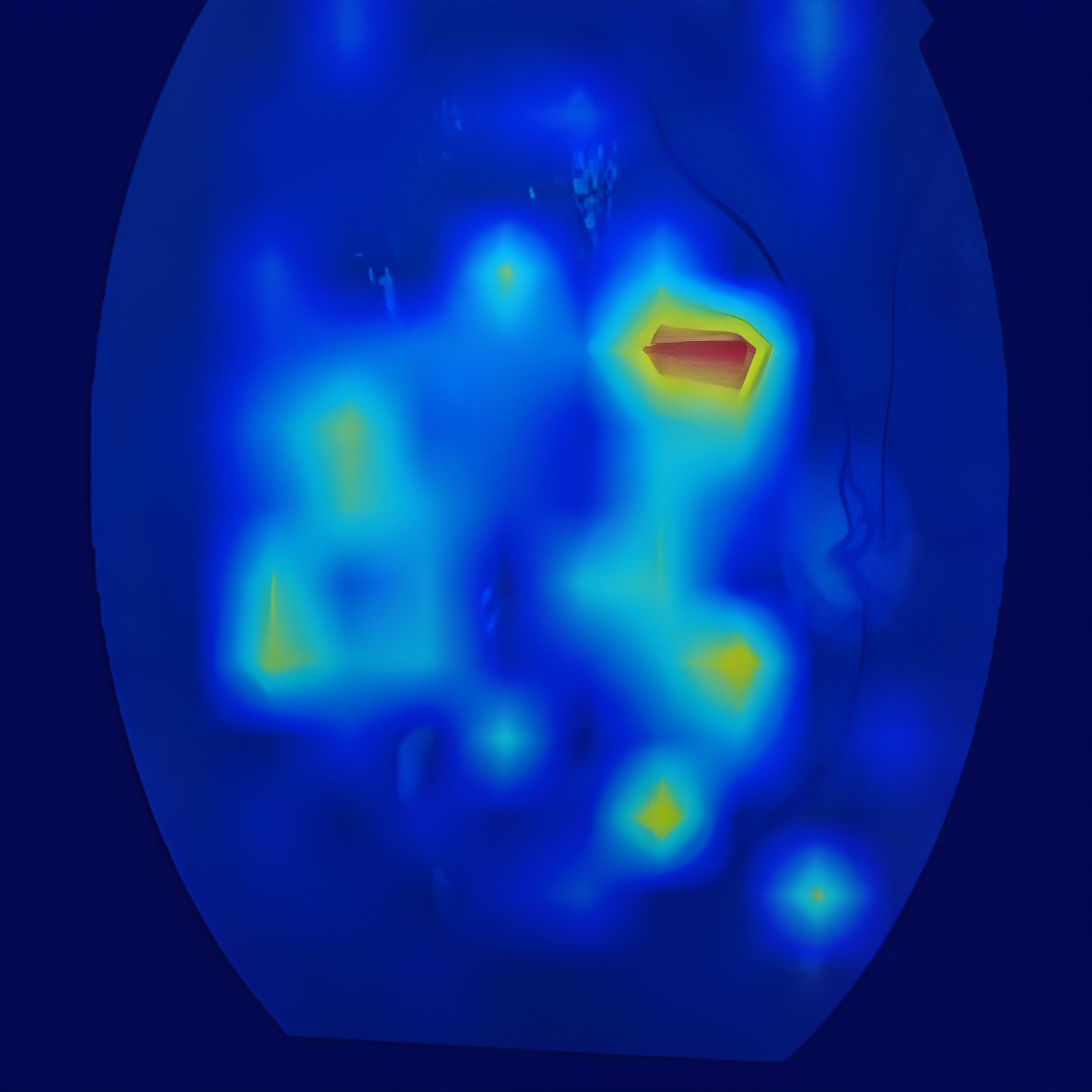} &
    \includegraphics[width=.0875\textwidth, height=.09\textwidth]{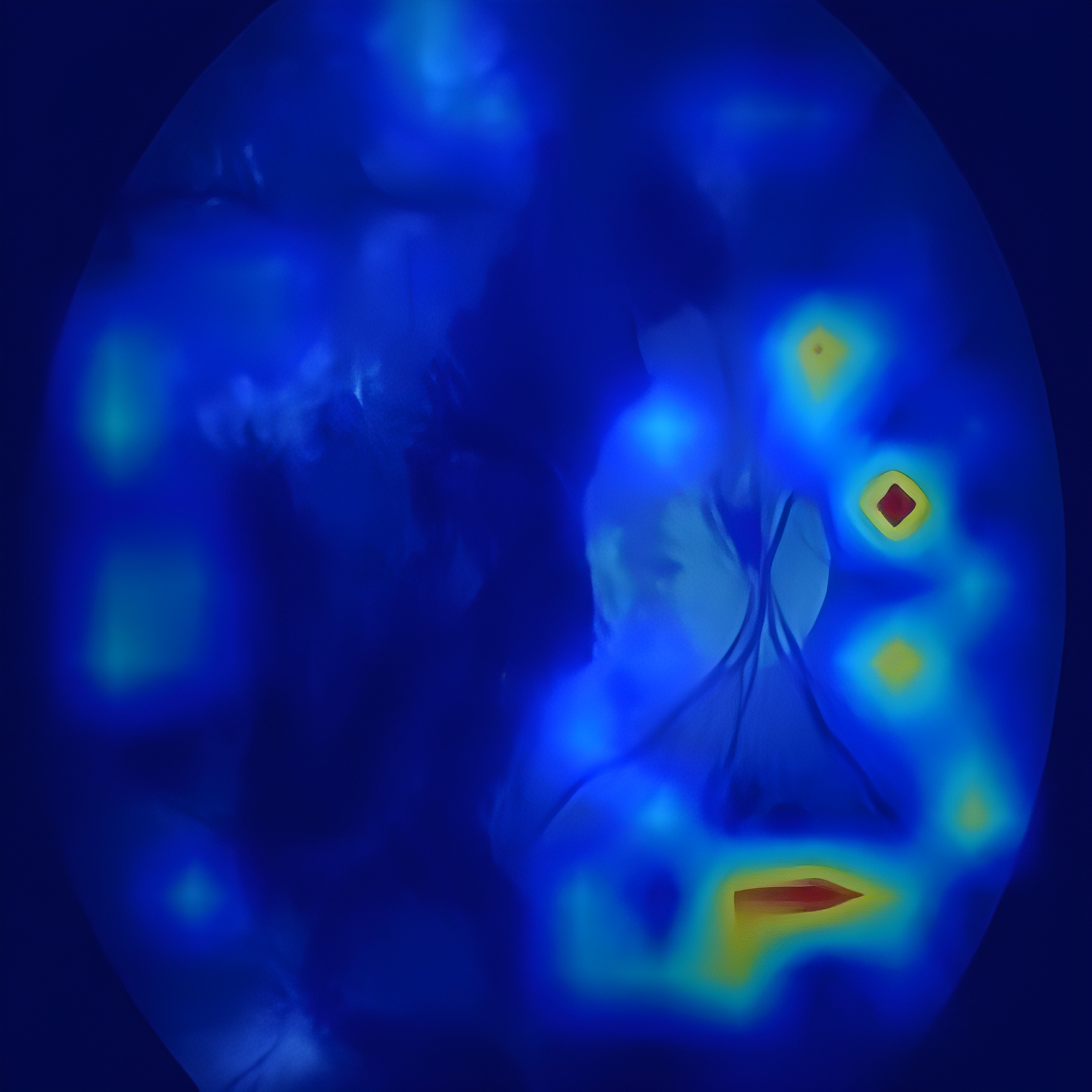}\\ \hline
& && && \\[\dimexpr-\normalbaselineskip+1.5pt]
\rotatebox{90}{~DeiT}    &{\includegraphics[width=.0875\textwidth, height=.09\textwidth]{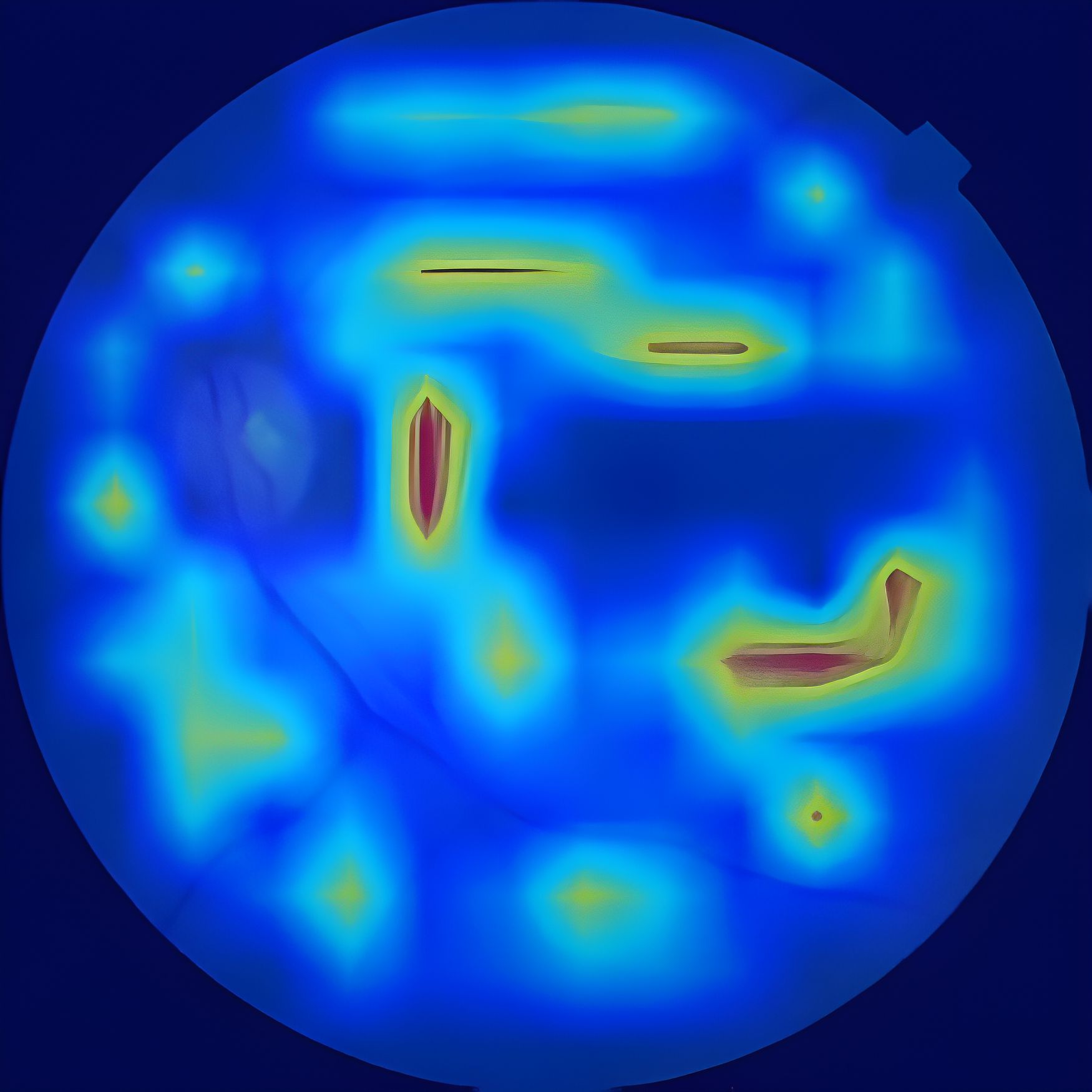}} &
    {\includegraphics[width=.0875\textwidth, height=.09\textwidth]{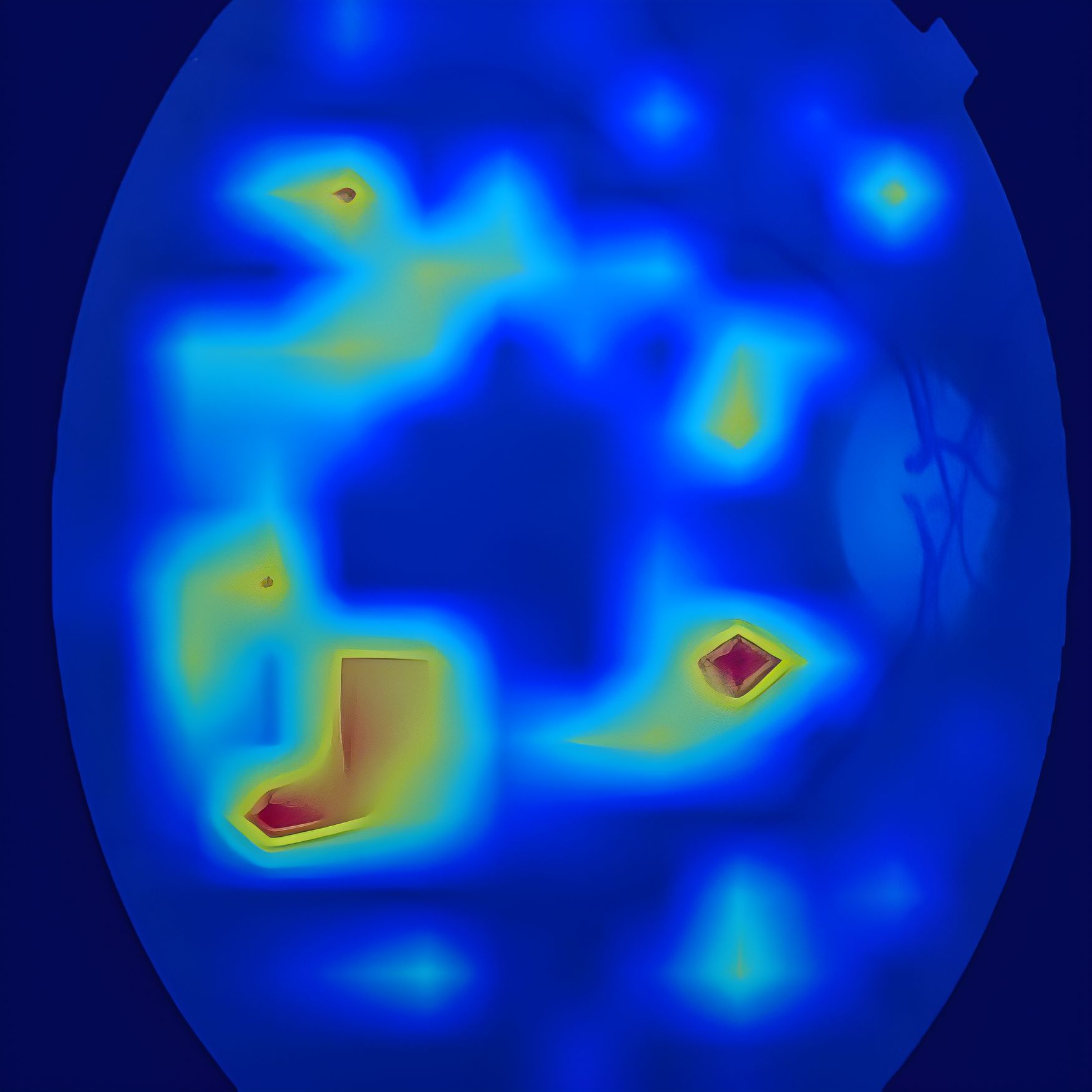}} &
    {\includegraphics[width=.0875\textwidth, height=.09\textwidth]{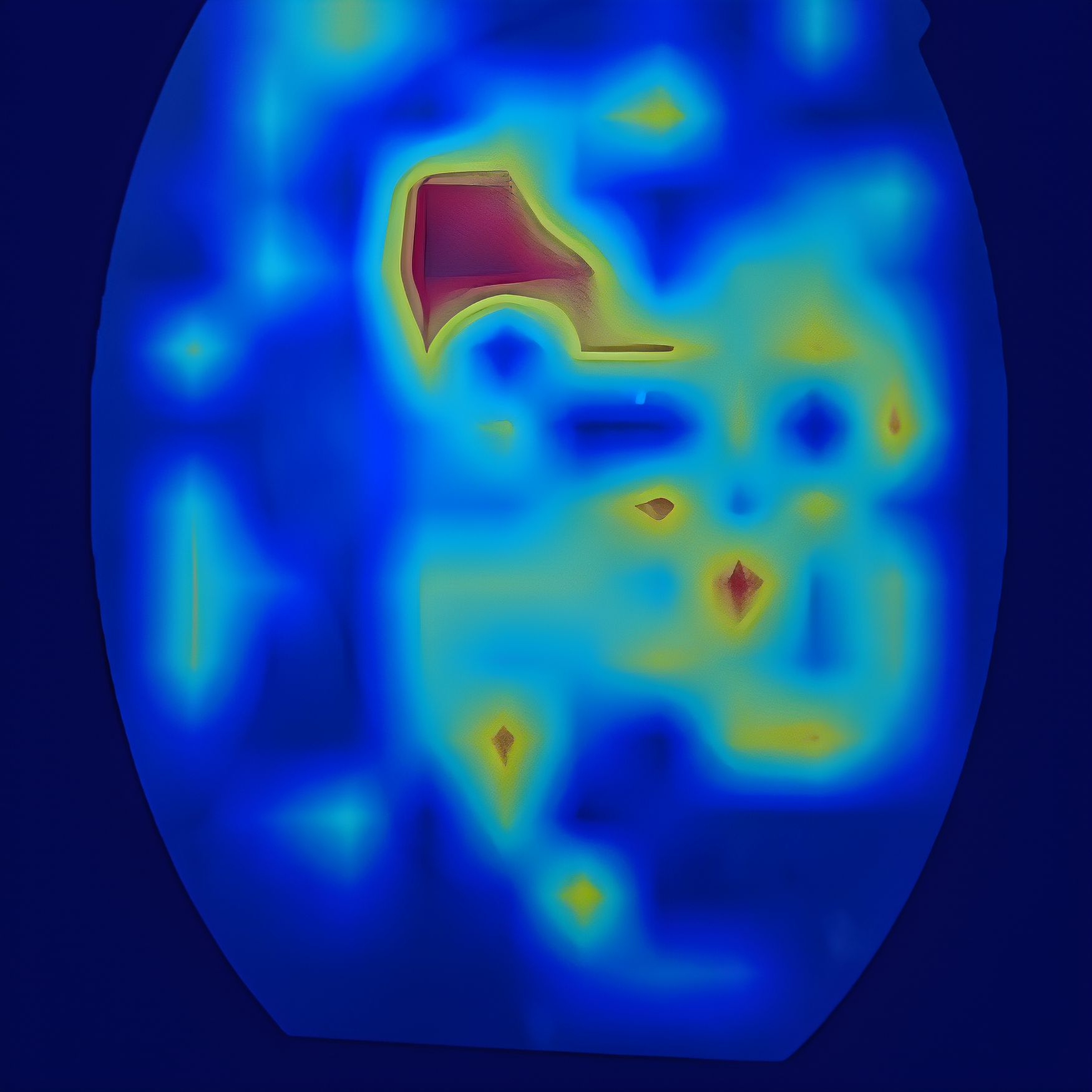}} &
    \includegraphics[width=.0875\textwidth, height=.09\textwidth]{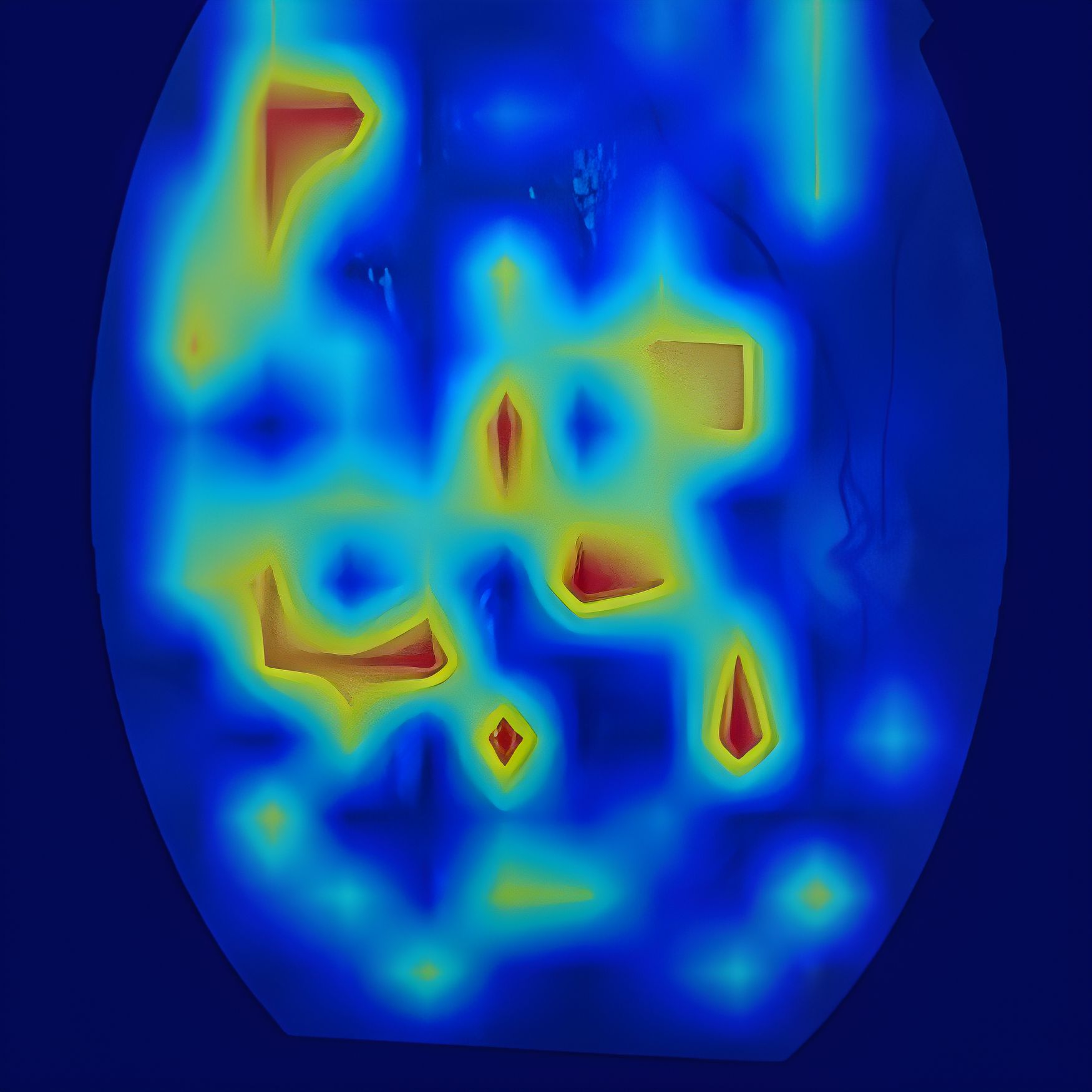} &
    \includegraphics[width=.0875\textwidth, height=.09\textwidth]{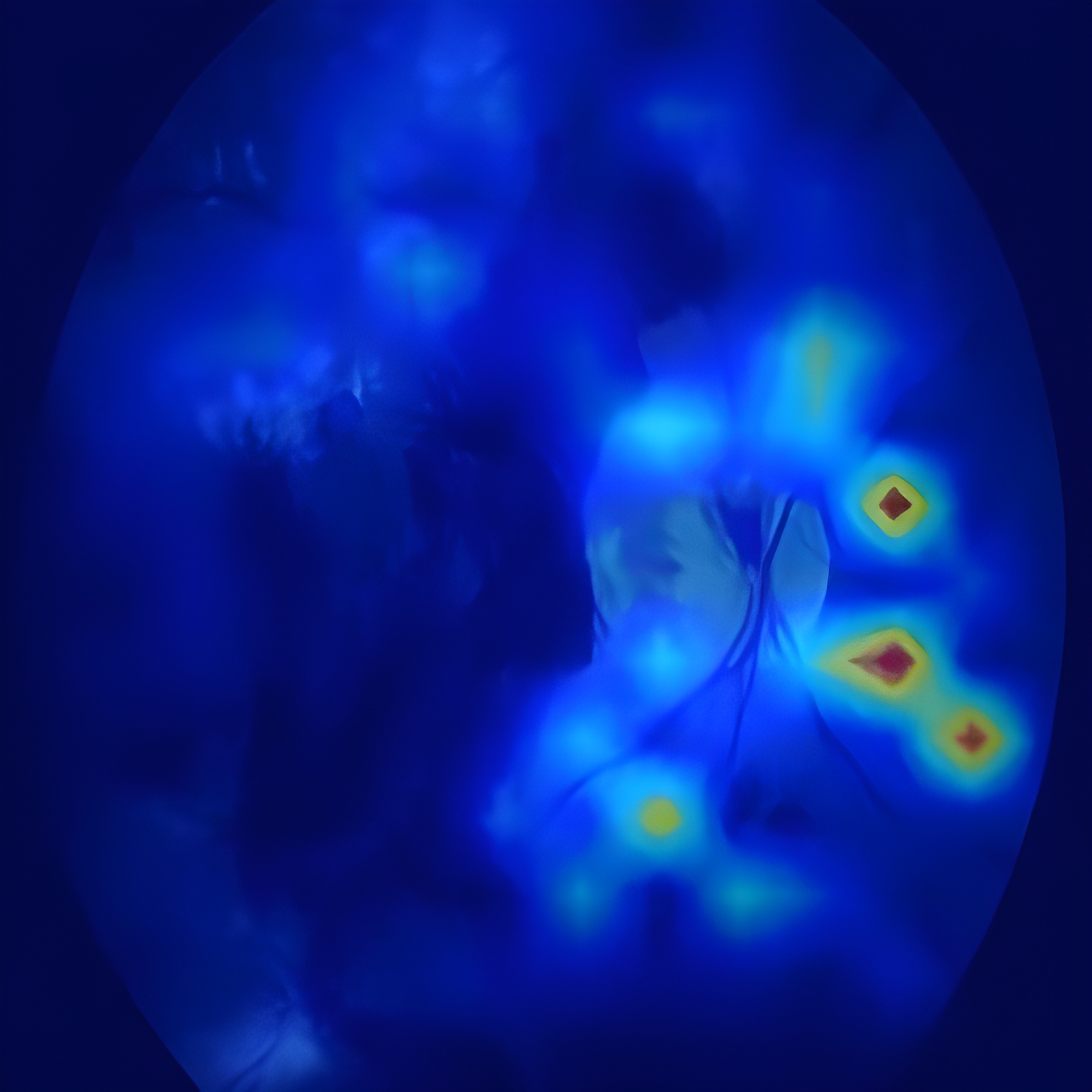}\\ \hline
& && &&  \\[\dimexpr-\normalbaselineskip+1.5pt]
\rotatebox{90}{~BEiT}    &{\includegraphics[width=.0875\textwidth, height=.09\textwidth]{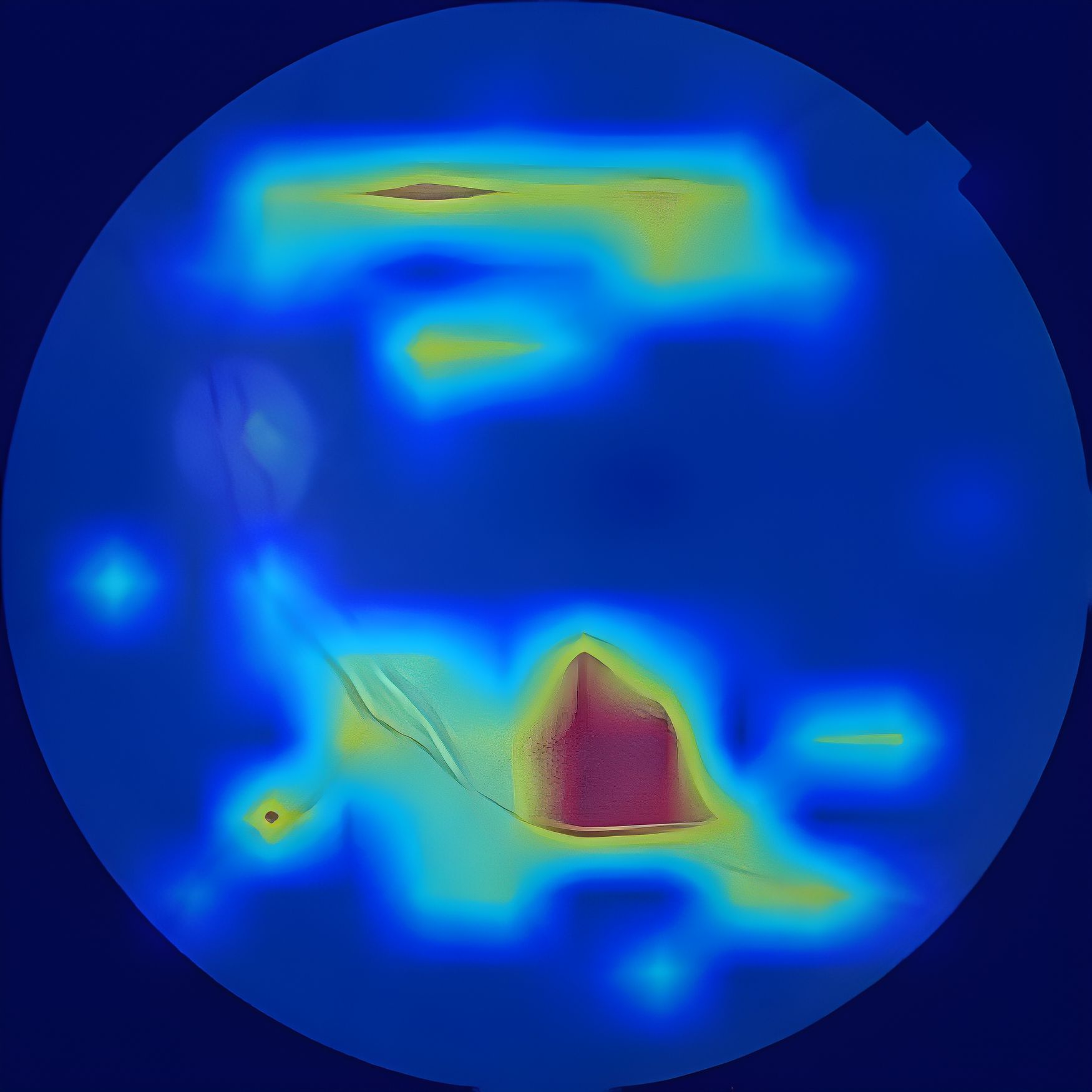}} &
    {\includegraphics[width=.0875\textwidth, height=.09\textwidth]{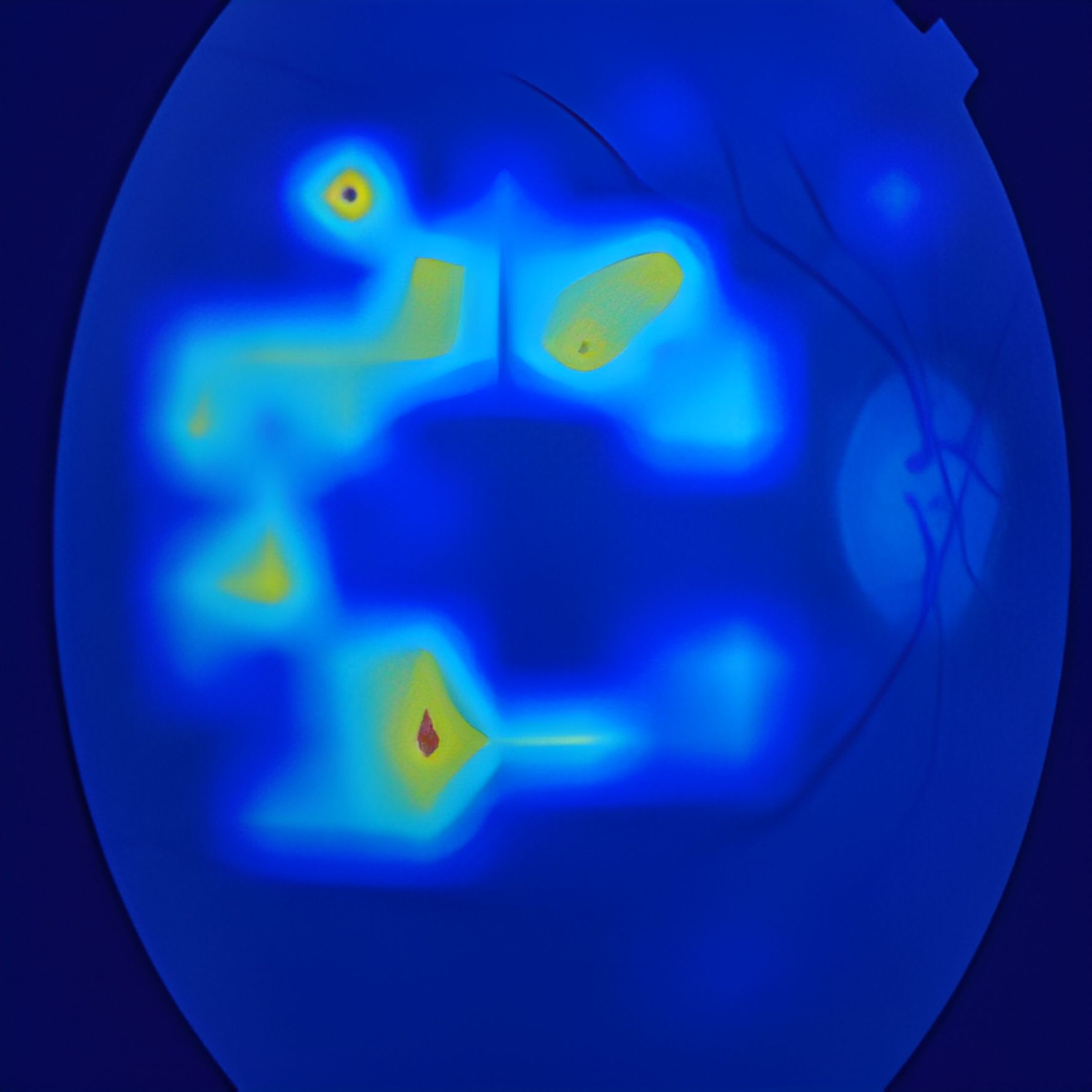}} &
    {\includegraphics[width=.0875\textwidth, height=.09\textwidth]{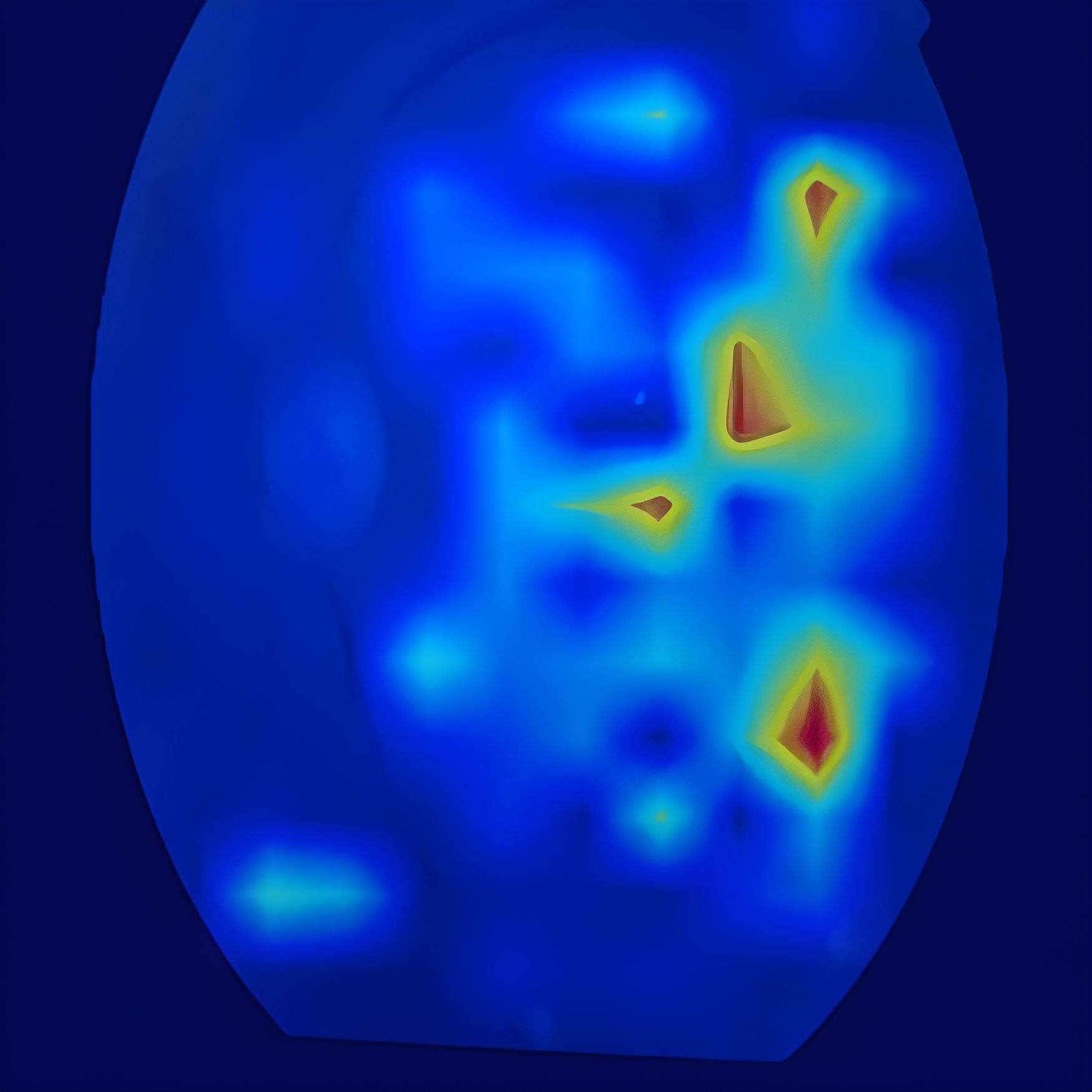}} &
    \includegraphics[width=.0875\textwidth, height=.09\textwidth]{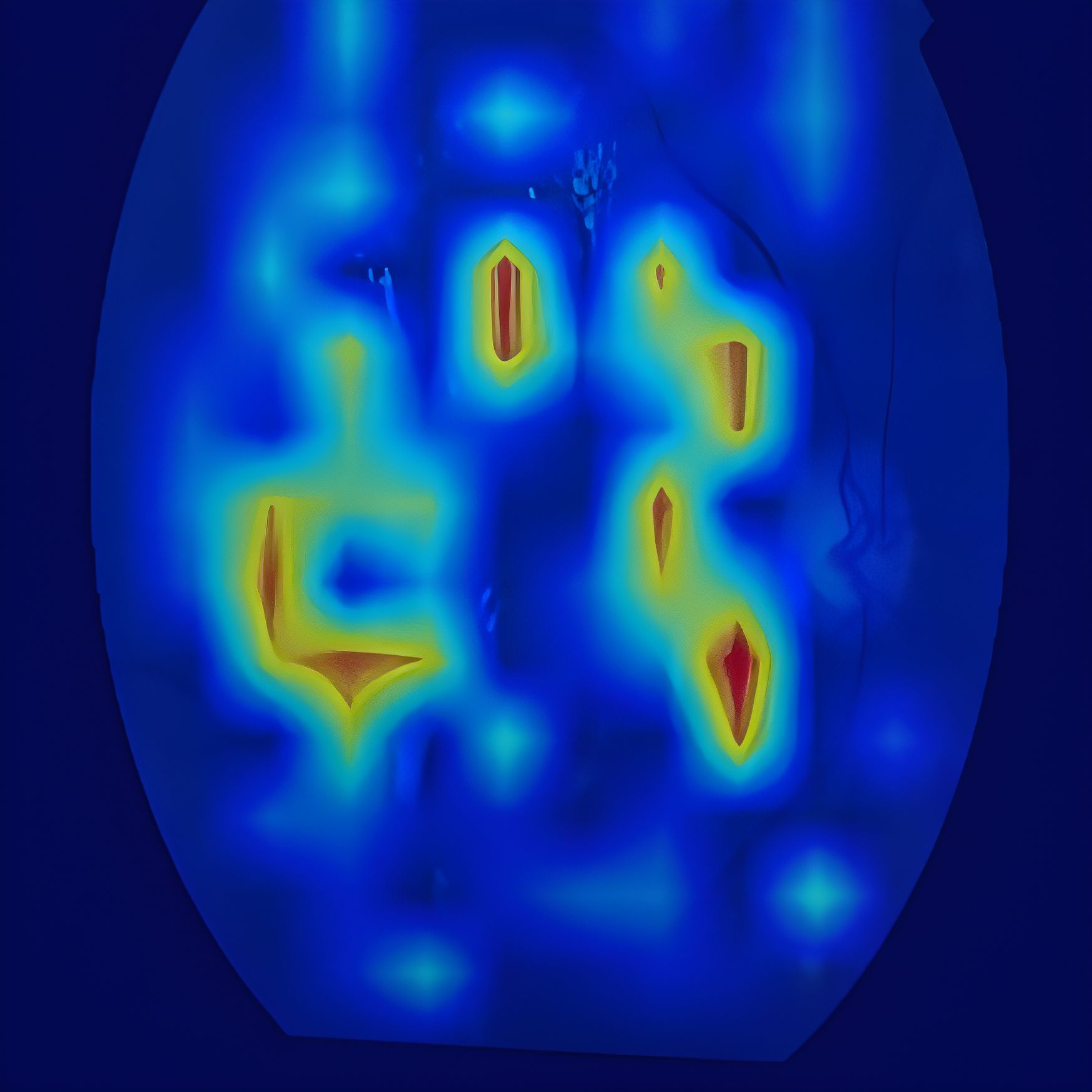} &
    \includegraphics[width=.0875\textwidth, height=.09\textwidth]{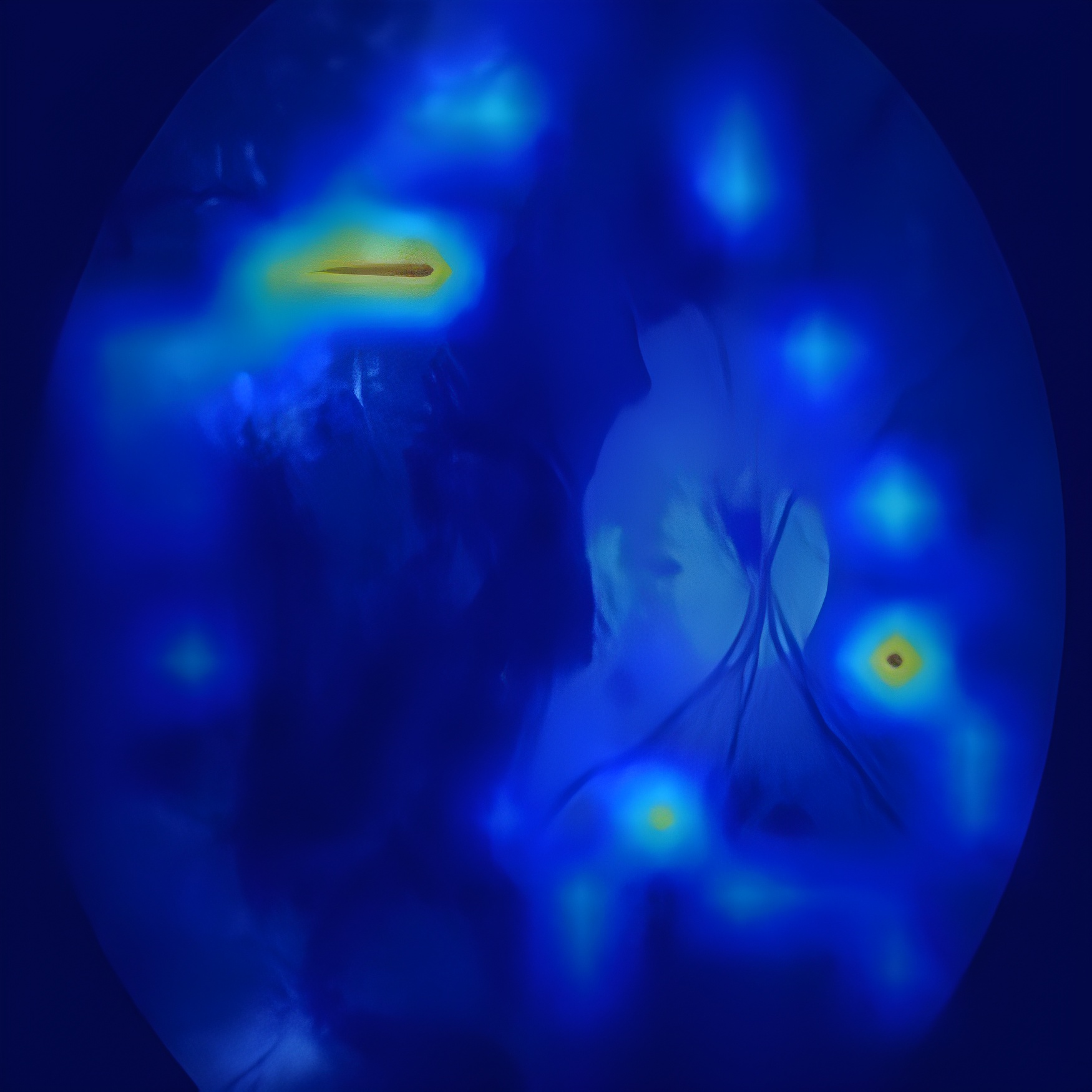}\\ \hline
& && && \\[\dimexpr-\normalbaselineskip+1.5pt]
\rotatebox{90}{~CaiT}    &{\includegraphics[width=.0875\textwidth, height=.09\textwidth]{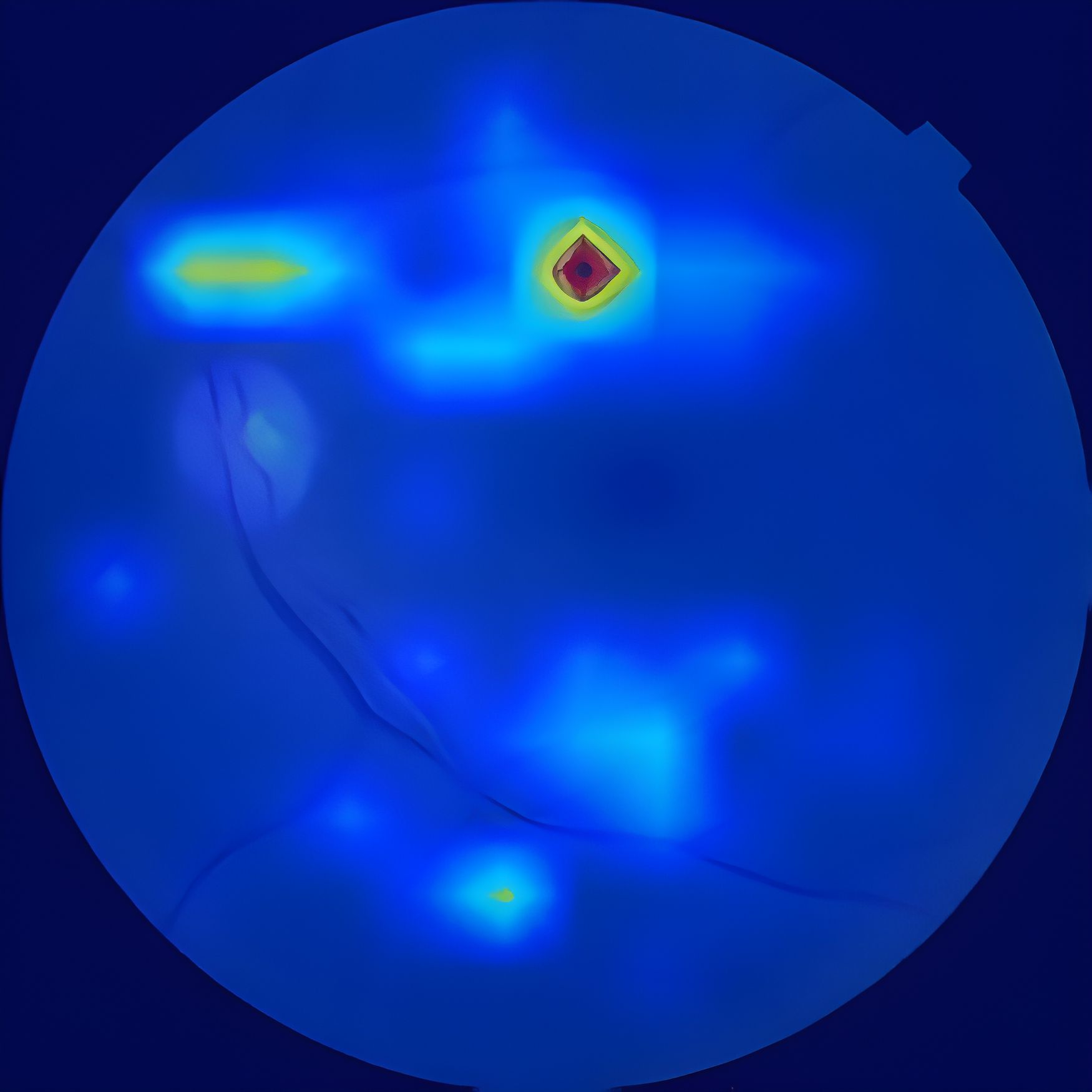}} &
    {\includegraphics[width=.0875\textwidth, height=.09\textwidth]{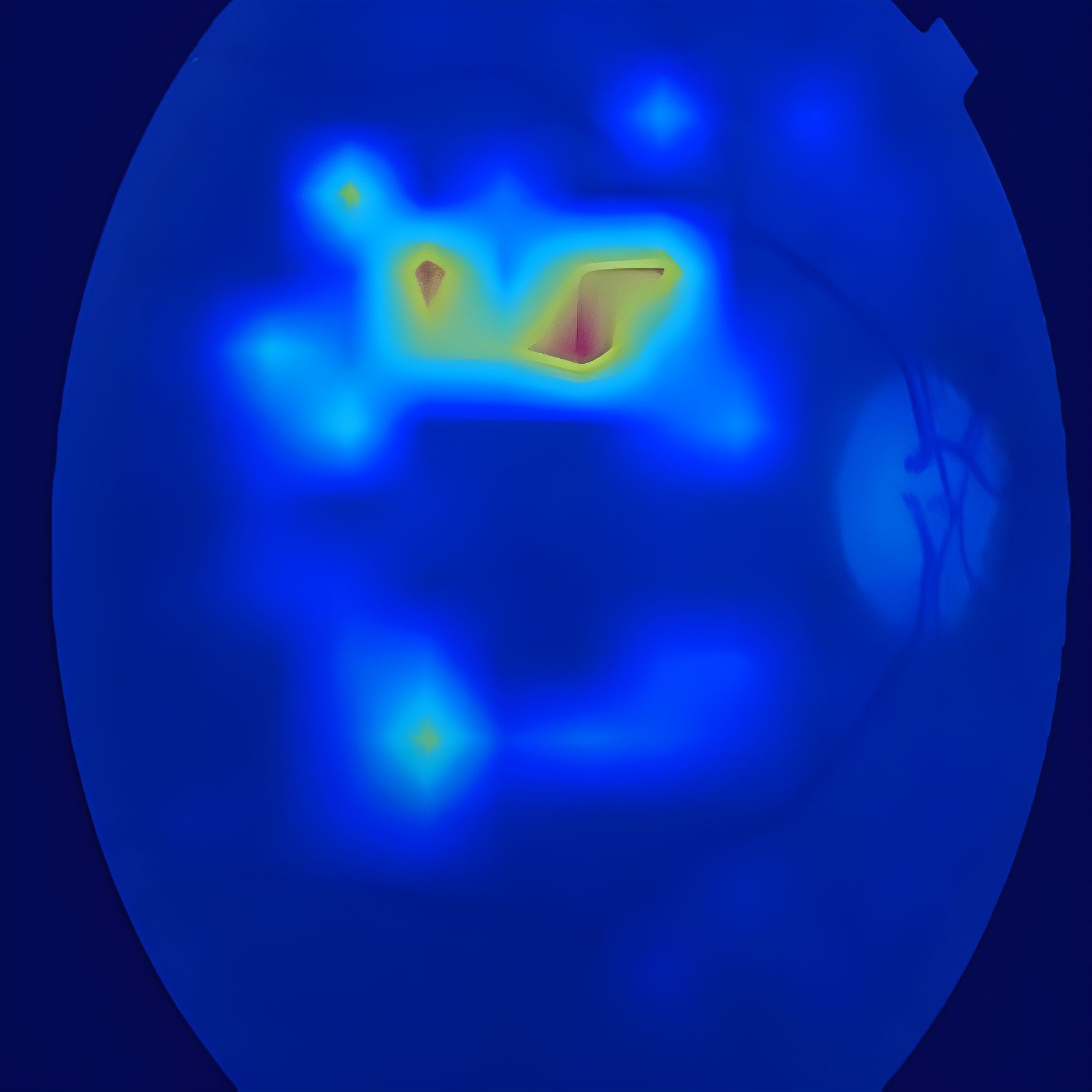}} &
    {\includegraphics[width=.0875\textwidth, height=.09\textwidth]{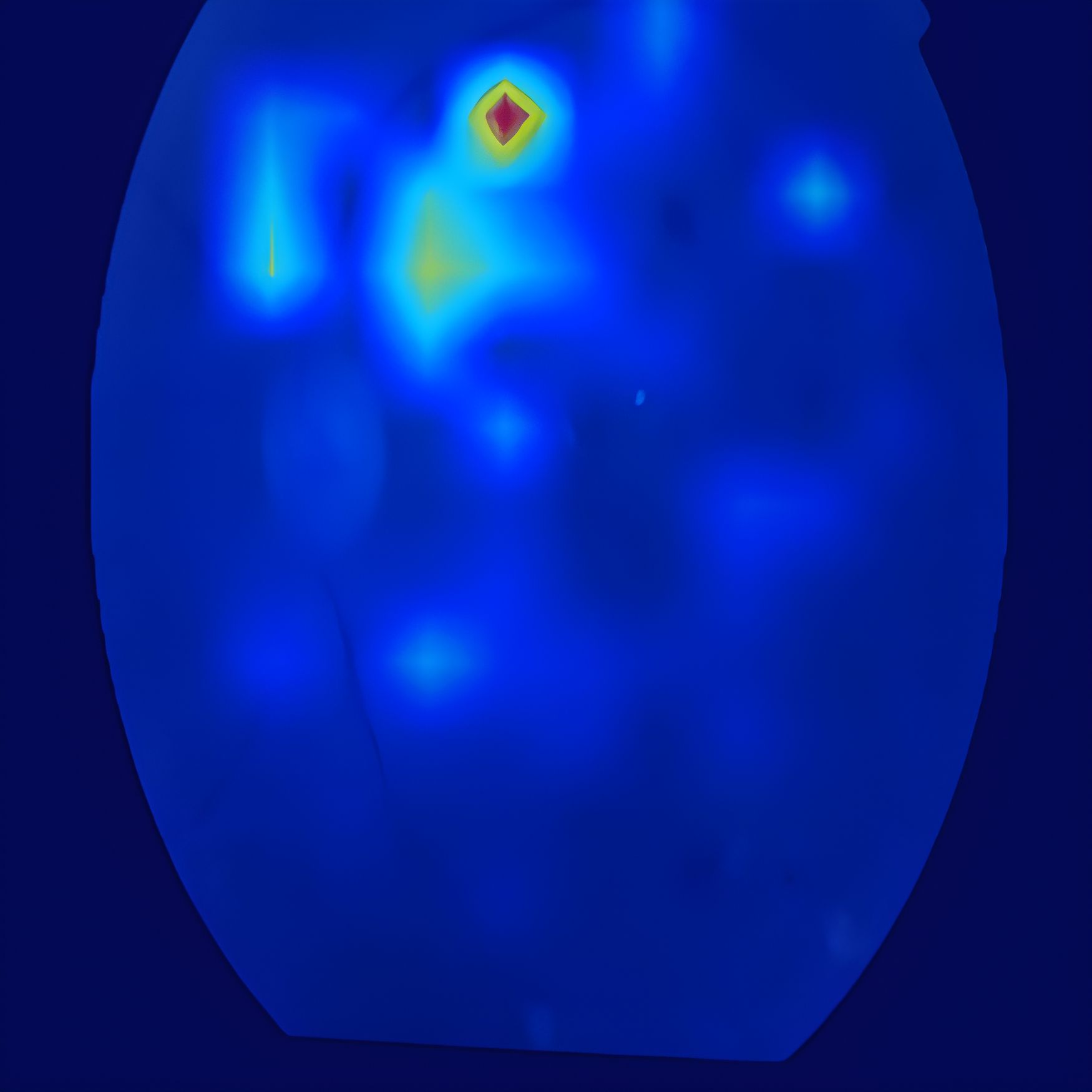}} &
    \includegraphics[width=.0875\textwidth, height=.09\textwidth]{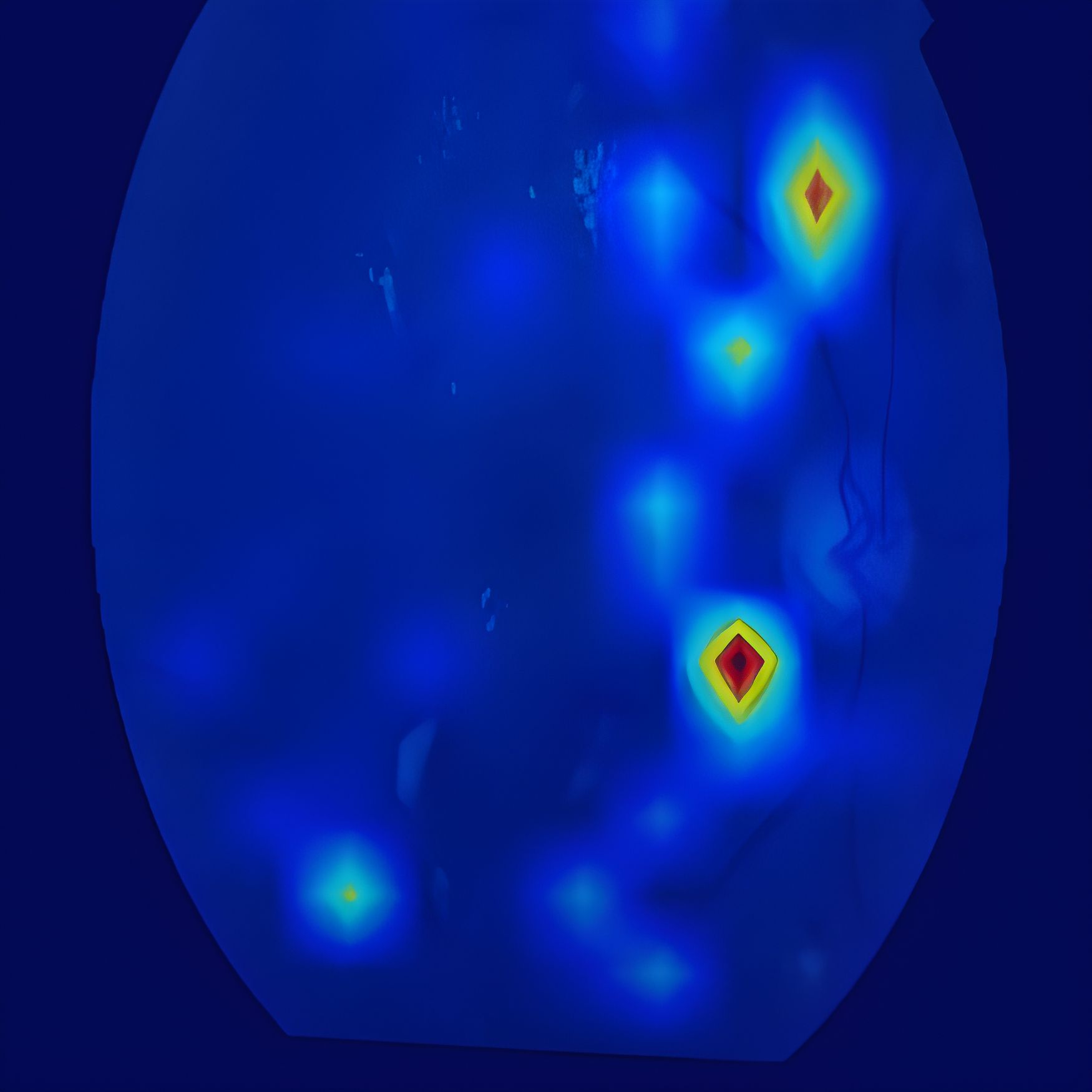} &
    \includegraphics[width=.0875\textwidth, height=.09\textwidth]{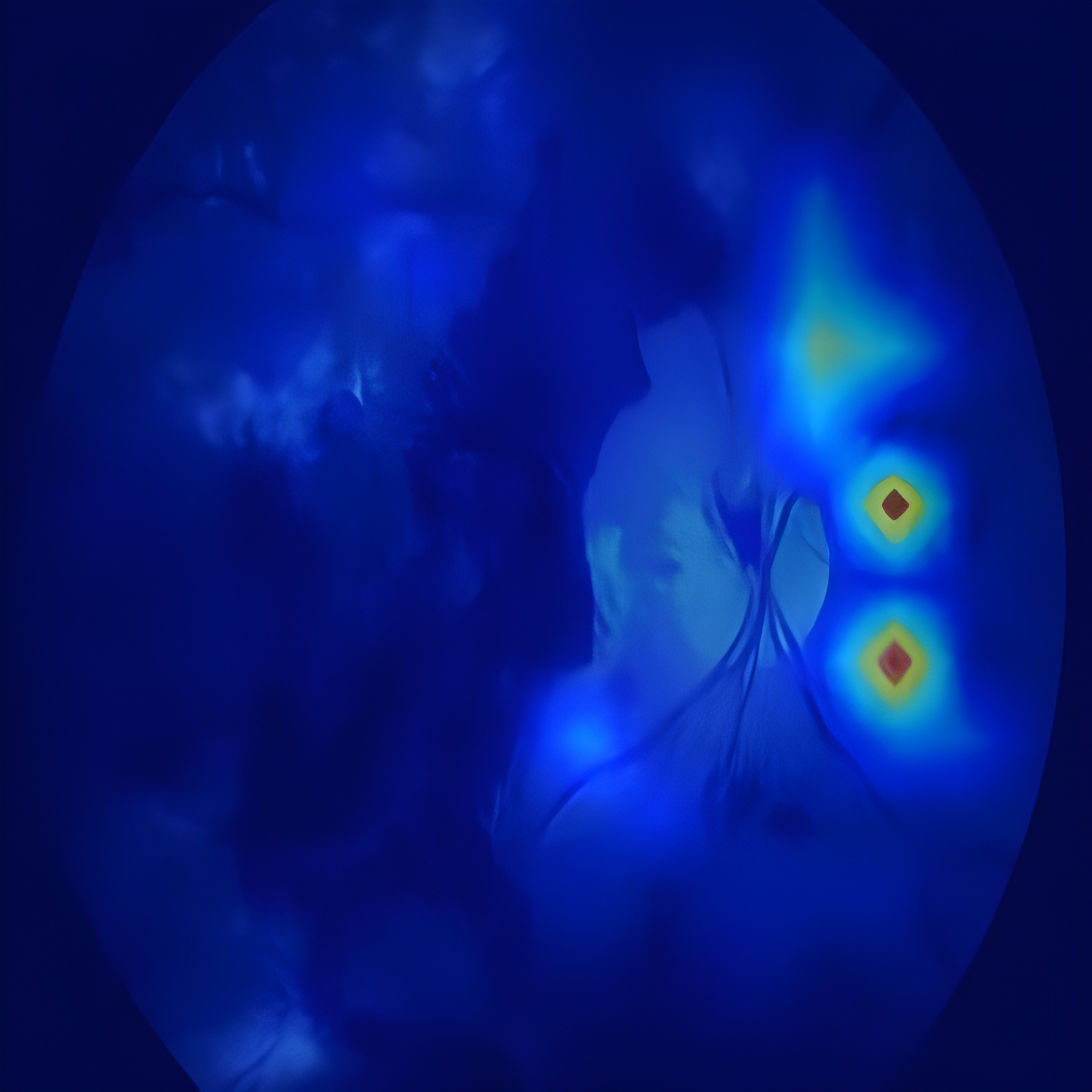}\\ \hline
& && && \\[\dimexpr-\normalbaselineskip+1.5pt]
%
& negative & mild & moderate & severe & proliferative
\end{tabular}
\end{adjustbox}
\caption{Fundus images (1$^{st}$ row) with Grad-CAM maps for ViT, DeiT, BEiT, CaiT as shown in 2$^{nd}$, 3$^{rd}$, 4$^{th}$, 5$^{th}$ rows, respectively}
\label{fig:gradcam}
\end{figure}

As mentioned in subsection \ref{subsubsec:model_performance}, we present the Grad-CAM maps of the employed individual image transformers in Fig. \ref{fig:gradcam}.

\section{Tuned Weighted-Mean Ensembled Transformer}
\label{app:tuned_weights}

\textcolor{black}{As mentioned in subsection \ref{subsubsec:hyperparam}, in Table \ref{tab:weights2}, we provide the tuned values of $\alpha_j$ that contributed to achieving the best performance of the transformers ensembled with the weighted mean scheme ($EiT_{wm}$).} 


\begin{table}[!ht]
\centering
\scriptsize
\caption{Tuned weights $\alpha_j$ of $EiT_{wm}$}
    \begin{tabular}{c|c|c|c|c}
    \hline
Transformers$_{wm}$ & $\alpha_1$ & $\alpha_2$ & $\alpha_3$ & $\alpha_4$ \\ \hline \hline
ViT + DeiT   & 0.25 & 0.75 & - & - \\ \hline
ViT + BEiT   & 0.4 & 0.6 & - & -\\ \hline
ViT + CaiT & 0.4 & 0.6 & - & -\\ \hline
DeiT + BEiT   & 0.4 & 0.6 & - & -\\ \hline
DeiT + CaiT & 0.3 & 0.7 & - & -\\ \hline
BEiT + CaiT & 0.5 & 0.5 & - & -\\ \hline
ViT + DeiT + BEiT   & 0.2 & 0.3 & 0.5 & -\\ \hline
ViT + DeiT + CaiT & 0.2 & 0.3 & 0.5 & -\\ \hline
ViT + BEiT + CaiT & 0.2 & 0.4 & 0.4 & -\\ \hline
DeiT + BEiT + CaiT & 0.3 & 0.3 & 0.4 & -\\ \hline
ViT + DeiT + BEiT + CaiT & 0.1 & 0.1 & 0.4 & 0.4\\ \hline
    \end{tabular}
    \label{tab:weights2}
\end{table}

\bibliographystyle{elsarticle-num-names} 
\bibliography{refs}

\end{document}